\documentclass[nohyperref]{article}

\usepackage{microtype}
\usepackage{graphicx}
\usepackage{subfigure}
\usepackage{booktabs} 

\usepackage{hyperref}
\usepackage{algorithm2e}

\usepackage[accepted]{icml2022}

\usepackage{amsmath}
\usepackage{amssymb}
\usepackage{mathtools}
\usepackage{amsthm}

\usepackage{xcolor}         
\usepackage[english]{babel}
\usepackage{graphicx}
\usepackage{verbatim}

\definecolor{olivegreen}{HTML}{006400}
\definecolor{echoblue}{HTML}{0099CC}
\definecolor{gold}{HTML}{d18600}
\definecolor{vividred}{HTML}{E60B42}
\definecolor{echonavy}{HTML}{0054B2}
\definecolor{darkgry}{HTML}{333333}
\definecolor{echopurple}{HTML}{9400D1}

\usepackage[capitalize,noabbrev]{cleveref}

\theoremstyle{plain}

\theoremstyle{definition}

\theoremstyle{remark}

\renewcommand{\vec}{\mathbf}

\usepackage[textsize=tiny]{todonotes}

\icmltitlerunning{The CLRS Algorithmic Reasoning Benchmark}

\begin{document}

\twocolumn[
\icmltitle{The CLRS Algorithmic Reasoning Benchmark}



\icmlsetsymbol{equal}{*}

\begin{icmlauthorlist}
\icmlauthor{Petar Veli\v{c}kovi\'{c}}{dm}
\icmlauthor{Adri\`{a} Puigdom\`{e}nech Badia}{dm}
\icmlauthor{David Budden}{dm}\\
\icmlauthor{Razvan Pascanu}{dm}
\icmlauthor{Andrea Banino}{dm}
\icmlauthor{Misha Dashevskiy}{dm}
\icmlauthor{Raia Hadsell}{dm}
\icmlauthor{Charles Blundell}{dm}
\end{icmlauthorlist}

\icmlaffiliation{dm}{DeepMind}

\icmlcorrespondingauthor{Petar Veli\v{c}kovi\'{c}}{petarv@deepmind.com}

\icmlkeywords{Machine Learning, ICML}

\vskip 0.3in
]



\printAffiliationsAndNotice{}  

\begin{abstract}
 Learning representations of algorithms is an emerging area of machine learning, seeking to bridge concepts from neural networks with classical algorithms. Several important works have investigated whether neural networks can effectively reason like algorithms, typically by learning to execute them. The common trend in the area, however, is to generate targeted kinds of algorithmic data to evaluate specific hypotheses, making results hard to transfer across publications, and increasing the barrier of entry. To consolidate progress and work towards unified evaluation, we propose the CLRS Algorithmic Reasoning Benchmark, covering classical algorithms from the Introduction to Algorithms textbook. Our benchmark spans a variety of algorithmic reasoning procedures, including sorting, searching, dynamic programming, graph algorithms, string algorithms and geometric algorithms. We perform extensive experiments to demonstrate how several popular algorithmic reasoning baselines perform on these tasks, and consequently, highlight links to several open challenges. Our library is readily available at \url{https://github.com/deepmind/clrs}.
\end{abstract}

\section{Introduction}

Neural networks and classical algorithms are two techniques that operate on diametrically opposite (and complementary) sides of problem-solving: neural networks can adapt and generalise to raw inputs, automatically extracting appropriate features and a single neural network setup is often applicable to many separate tasks \citep{zamir2018taskonomy}. However, they are hard to interpret, notoriously unreliable when extrapolating outside of the dataset they have been trained on, and rely on massive quantities of training data. On the other hand, algorithms trivially strongly generalise to inputs of arbitrary sizes, and can be verified or proven to be correct, with interpretable step-wise operations. Their shortcoming is that inputs must be made to conform to a particular algorithm specification, and looking at a separate task often requires coming up with an entirely new algorithm \citep{velivckovic2021neural}.

Bringing the two sides closer together can therefore yield the kinds of improvements to performance, generalisation and interpretability that are unlikely to occur through architectural gains alone. Accordingly, algorithmic modelling as a domain for testing neural networks has been gaining popularity over the last few years~\citep{zaremba2014learning,kaiser2015neural,trask2018neural,vinyals2015pointer,kool2018attention,freivalds2019neural,dwivedi2020benchmarking,chen2020can,tang2020towards,velivckovic2019neural,yan2020neural,deac2020graph} due to its ability to highlight various reasoning limitations of existing architectures. 

Earlier work \citep{zaremba2014learning,kaiser2015neural} focused on the need of long-term memory capabilities when executing algorithms, which offered a good test-bed for various recurrent and memory architectures. Recently, algorithmic tasks have been used to highlight the efficiency of graph neural networks~\citep{dwivedi2020benchmarking,chen2020can,velivckovic2019neural,yan2020neural,corso2020principal,tang2020towards,georgiev2020neural,velivckovic2020pointer} and to distinguish between different variations of them, typically through the lens of \emph{algorithmic alignment}---architectures that align better with the underlying algorithm can be proven to have better sample complexity \citep{xu2019can}. Unfortunately, many of these works remain disconnected in terms of the algorithms they target, how the data is presented to the model or through the training and testing protocols they use, making direct comparison somewhat difficult.

To make a first step towards a unified benchmark for algorithmic reasoning tasks, we propose a comprehensive dataset which we will refer to as \emph{The CLRS Algorithmic Reasoning Benchmark}, in homage to the \emph{Introduction to Algorithms} textbook by Cormen, Leiserson, Rivest and Stein \citep{cormen2009introduction}.
                    
\begin{figure}[ht!]
\centering
\includegraphics[trim=20 600 0 0, clip,width=\linewidth]{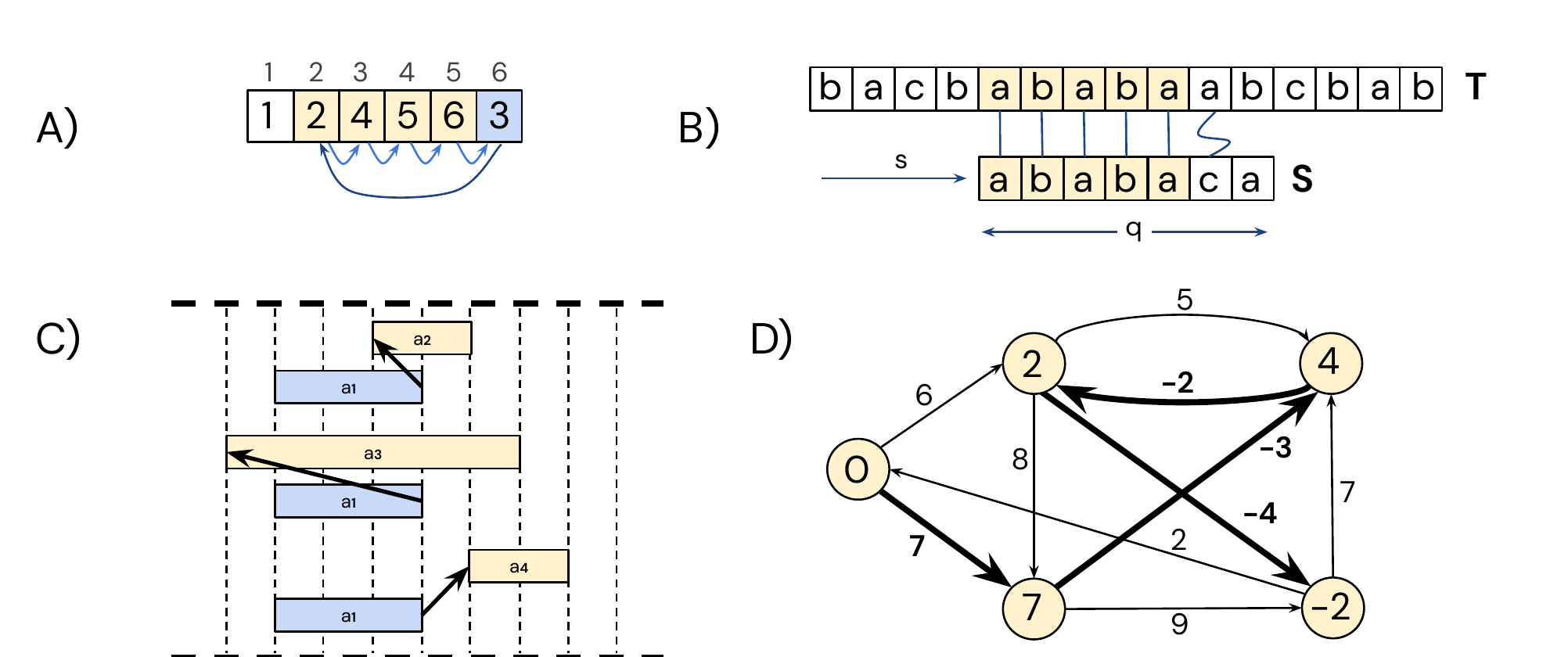}
\caption{Example of four algorithms within CLRS-30. A) insertion sort; B) string matching; C) greedy task scheduling; D) shortest paths.}
\label{fig:CLRS-B}
\end{figure}

Within this benchmark, we propose and evaluate on \textbf{CLRS-30}: a dataset containing trajectories---a trajectory is formed of inputs, the corresponding outputs and optional intermediary targets---of 30 classical algorithms covering various forms of reasoning, including sorting, searching, dynamic programming, geometry, graphs and strings. Some of these algorithms are depicted in Figure~\ref{fig:CLRS-B}. The appeal and motivation for such a benchmark goes beyond unifying or providing a common ground for previous works, as we will describe. We believe that CLRS-30 is well positioned to explore \emph{out-of-distribution} (OOD) generalization and transfer (as potentially part of a meta-learning setting) given the explicit and known relationship between different algorithms (e.g. what subroutines are shared and so forth). 

\section{Motivation}

Timely posed benchmarks have led to a significant progress in the field, from the impact of ImageNet \citep{ILSVRC15} on the vision community, to that of Wikipedia and Penn Treebank in popularizing neural networks for language modelling \citep{merity2016pointer, Mikolov11} or Atari-2600 for deep reinforcement learning \citep{bellemare2013arcade}. The prevalence of recent works focusing on algorithmic reasoning\footnote{Concurrent works published at the same venue include: \cite{xu2019can,velivckovic2019neural} at ICLR'20 and \cite{velivckovic2020pointer,corso2020principal,tang2020towards} at NeurIPS'20.}, as well as a history of disparate work on a variety of bespoke benchmarks~\cite{graves2014neural, zaremba2014learning, kaiser2015neural, trask2018neural}, suggests significant utility in a benchmark covering a wide-range of classical CS algorithms.

Learning to mimic an algorithm also provides an opportunity to extensively test the limitations of architectures both in terms of their representation capacity and processing. This can then be related back directly onto underlying operations and qualities of the well-studied CS algorithms being mimicked as we are aware of both the process used to generate the inputs and the specifics of the underlying function producing the corresponding outputs. Hence, benchmarking in this area can be used to better understand the limitations of current architectures and the optimisation schemes used. This benchmarking can come in many forms:

Data can be easily generated, allowing the neural network behaviour to be probed under different regimes: from few-shot learning all the way to infinite-data.

Algorithms can be used to understand the efficiency of different inductive biases and neural components. For example, a recent study \citep{tang2020towards} has demonstrated the direct benefits of choosing inductive biases that align well with iterative algorithms. Algorithms have also been used to highlight the importance of attention mechanisms \citep{graves2014neural} or to disambiguate various message passing mechanisms for graph neural networks \citep{richter2020normalized,joshi2020learning,velivckovic2019neural}.

Algorithms can require repeated computation, recursion, or performing very different forms of computations conditioned on the input, providing an excellent test-bed for evaluating compositionality; i.e. whether an algorithm executor can effectively exploit these repeated computations. 

One can control the amount of memory required to solve a problem instance, hence test the memorization ability of neural networks. Moreover, one can build a curriculum of tasks of increasing memory requirements~\citep{zaremba2014learning}.

Control over the difficulty of problem instances also allows the behaviour of a trained model to be tested on OOD samples. While neural networks are highly efficient on solving complex perceptual tasks, current theoretical understanding suggests that their power relies on their ability to \emph{interpolate}~\citep{Liu2020TheoryDL,belkin2019two, Jacot2018NTK}, limiting them to \emph{in-distribution} generalisation. General reasoning systems, however, need to be able to expand beyond this type of generalization. \emph{OOD generalization} \citep{li2020strong} is paramount, as generally one can not control the distribution a model will face over time when deployed. 

Understanding how algorithms operate on corner cases is a standard approach for analysing their correctness. Similarly, understanding the behaviour of a trained model on larger instances of the problem, or instances that expose such corner cases that were not covered in the training set, can elucidate to what degree the model has truly learned the \emph{algorithm} (as opposed to overfitting to specific statistics of the training data). Particularly, we can control how far from the training distribution a test instance is, potentially allowing us to understand to what extent the model generalizes OOD, and under which circumstances. In turn, this can offer insight into the effectiveness of different inductive biases, highlighting what kinds of inductive biases are useful for mimicking reasoning processes.

One would also expect a general reasoning system to be able to \emph{reuse} parts of learned computations when learning a new task, and to \emph{compose} learnt computational subroutines~\cite{Lake19,Griffiths19,alet18a}. These forms of generalization have been the aim of several learning paradigms from transfer learning to meta-learning and continual learning or domain adaptation. However, many of these paradigms rely on the concept of a \emph{task}, and measuring or understanding the ability of a learned system to \emph{reuse} or \emph{compose} requires the ability to decompose a task into sub-tasks and to be able to relate tasks among themselves. In many scenarios, such decompositions are ambiguous. Without a clear segmentation into sub-tasks, there can be no clearly defined distance metric between tasks~\cite{du2018adapting}. Conversely, algorithms are built based on subroutines that tend to be extensively shared, providing a good playground for formalizing and measuring \emph{reuse} and \emph{composition}, making an algorithmic reasoning benchmark potentially attractive to meta-learning practitioners.

Lastly and fundamentally, computer scientists rely on a relatively small\footnote{The entire Introduction to Algorithms textbook \citep{cormen2009introduction} proposes and discusses $\sim$100 algorithms in total.} number of algorithms to address an extremely vast set of problems. They can be seen as a very powerful basis that spans most forms of reasoning processes. On one hand, this means that any generic reasoning system likely has to be able to reproduce all such kinds of procedures, hence, building a system that properly learns all of them is a major stepping stone towards generic reasoning. On the other hand, this means that they can be used to discover inductive biases that will enable tackling more complex problems. This is either because these complex problems can be seen as a combination of several algorithms, or because learning certain algorithms can provide a reliable way for the model to learn how to access its own memory or how to attend to its input or other such internal mechanisms. So by first training on algorithms---potentially controlling the difficulty of training instances---one can pre-train for tasks where full trajectories may not be available \citep{velivckovic2021reasoning}. One such example is discovering novel polynomial-time heuristics for combinatorial optimisation \citep{bengio2020machine,cappart2021combinatorial,khalil2017learning} or reinforcement learning \citep{deac2020xlvin}. Note that our focus with this benchmark lies in learning the basic algorithms themselves only--this in itself proves sufficiently challenging for neural networks, and is itself a useful outcome for the reasons highlighted above. However, we speculate that once a neural network can learn not only individual algorithms but novel combinations of multiple algorithms or even discover new algorithms, such networks will be useful in a wide variety of problems from scientific problems such as protein folding and genomics to simulated environments such as those used by reinforcement learning and control--much as classic CS algorithms already make in-roads into these domains but lack the ability to learn from data.

Guided by these observations, we regard CLRS-30 as a first step towards a pragmatic setting to test many of these different aspects of current architectures. While we do not directly target all of the scenarios outlined above, the benchmark was built with ease of expansion in mind; enabling for extensive tweaking of training/testing setups, kinds of information captured in algorithm trajectories, as well as including additional algorithms, which we aim to do consistently over time.

\section{CLRS Algorithmic Reasoning Benchmark}
\label{sec:CLRS3}

Owing to its name, CLRS-30 consists only of algorithms which may be encountered in the CLRS textbook \citep{cormen2009introduction}. Further, all algorithm trajectories and relevant variables have been designed to match the pseudocode in the textbook as closely as possible. We begin by describing the selection criteria we applied when determining which algorithms to include in CLRS-30.

Our initial survey of the textbook yielded 94 algorithms and data structures of interest. From this point, we set out to filter this set to algorithms suitable for inclusion in the initial version of our benchmark. The criteria we applied, with justification and remarks, are as follows: 

We want to be able to reliably generate \emph{ground-truth} outputs for large inputs. As such, NP-hard tasks (and approximation algorithms thereof) have been excluded. Our decision is backed up by theoretical work suggesting impossibility of accurately modelling NP-hard problems using polynomial-time samplers, unless NP$=$co-NP \citep{yehuda2020s}.

Tasks requiring \emph{numerical outputs} have been excluded. Evaluating their performance is ambiguous, and may be dependent on the way architectures choose to represent numbers. For example, \citet{yan2020neural} (which represents numbers in binary) and \citet{velivckovic2019neural} (which represents them in floating-point) report different metrics on predicting shortest-path lengths. This excludes most number-theoretic algorithms, linear programming, and max-flow\footnote{It should be noted that, by the max-flow min-cut theorem \citep{ford2015flows}, any max-flow problem can be cast as finding the minimum cut containing the source vertex. This is a discrete decision problem over input vertices, which hence doesn't violate our constraints, and could be included in future iterations.}. It does \emph{not} exclude shortest-path algorithms: we can treat them as tasks of finding edges belonging to the shortest path, as was done in \citet{velivckovic2019neural,tang2020towards}. The numerical values of path lengths are then treated as intermediate parts of the trajectory, and not directly evaluated on.

Standalone \emph{data structures} do not directly represent a task\footnote{In programming language terms, their algorithms tend to be of the \texttt{void} type.}. Rather, their target is appropriately updating the internal state of the data structure. Hence, we don't include their operations, unless they appear as components of algorithms. We, of course, look forward to including them in subsequent versions of the dataset, as they can provide useful building blocks for learning complex algorithms.

Lastly, there are representational issues associated with dynamically allocated memory---it may be unclear what is the best way to represent the internal memory storage and its usage in algorithm trajectories. One example of the ambiguity is in asking whether the algorithm executor should start with a ``scratch space'' defined by the space complexity of the problem that gets filled up, or dynamically generate such space\footnote{Akin to \texttt{malloc}-like calls in C++.} \citep{strathmann2021persistent}. As such, we for now exclude all algorithms that require allocating memory which cannot be directly attached to the set of objects provided at input time. This excludes algorithms like merge sort, Hierholzer's algorithm for finding Euler tours \citep{hierholzer1873moglichkeit}, or string matching using finite automata.

All of the above applied, we arrive at the 30 algorithms that are selected into CLRS-30, which we categorize as follows:

{\bf Sorting:} Insertion sort, bubble sort, heapsort \citep{williams1964algorithm}, quicksort \citep{hoare1962quicksort}.

{\bf Searching:} Minimum, binary search, quickselect \citep{hoare1961algorithm}.

{\bf Divide and Conquer (D\&C):} Maximum subarray (Kadane's variant \citep{bentley1984programming}).

{\bf Greedy:} Activity selection \citep{gavril1972algorithms}, task scheduling \citep{lawler1985traveling}.

{\bf Dynamic Programming:} Matrix chain multiplication, longest common subsequence, optimal binary search tree \citep{aho1974design}.

{\bf Graphs:} Depth-first and breadth-first search \citep{moore1959shortest}, topological sorting \citep{knuth1973fundamental}, articulation points, bridges, Kosaraju's strongly-connected components algorithm \citep{aho1974design}, Kruskal's and Prim's algorithms for minimum spanning trees \citep{kruskal1956shortest,prim1957shortest}, Bellman-Ford and Dijkstra's algorithms for single-source shortest paths \citep{bellman1958routing,dijkstra1959note} (+ directed acyclic graphs version), Floyd-Warshall algorithm for all-pairs shortest paths \citep{floyd1962algorithm}.

{\bf Strings:} Na\"{i}ve string matching, Knuth-Morris-Pratt (KMP) string matcher \citep{knuth1977fast}.

{\bf Geometry:} Segment intersection, Convex hull algorithms: Graham scan \citep{graham1972efficient}, Jarvis' march \citep{jarvis1973identification}.

The chosen algorithms span a wide variety of reasoning procedures, and hence can serve as a good basis for algorithmic reasoning evaluation, as well as extrapolation to more challenging problems.

\subsection{Implementation, probes and representation}\label{sect:impl}

We have implemented the selected 30 algorithms in an idiomatic way, which aligns as closely as possible to the original pseudocode from \citet{cormen2009introduction}. This allows us to automatically generate input/output pairs for all of them, enabling full control over the input data distribution, so long as it conforms to the preconditions of the algorithm. Further, we capture the intermediate algorithm trajectory in the form of \textbf{``hints''} (detailed in section~\ref{sec:hints}), which allow insight into the inner workings of the algorithm. Such trajectories have already been extensively used in related work \citep{velivckovic2019neural,velivckovic2020pointer,georgiev2020neural,deac2020graph} and are typically crucial for OOD generalisation.

In the most generic sense, algorithms can be seen as manipulating sets of objects, along with any relations between them (which can themselves be decomposed into binary relations). If the sets are (partially) ordered (e.g. arrays or rooted trees), this can be imposed by including predecessor links. Therefore, algorithms generally operate over \emph{graphs}. Motivated by existing theoretical results showing that graph neural networks align well with dynamic programming-style computations \citep{xu2019can,dudzik2022graph}, we propose a graph-oriented way to encode the data.

Generally, our data is represented as a set of $n$ vertices\footnote{Edges are only present to represent the predecessor vertex if the input is a partially ordered.}, where $n$ is a hyperparameter that is provided as part of the dataset generation process.

When the semantics of these nodes are not immediately clear from the task (e.g. graph algorithms naturally operate over a graph of $n$ nodes), we make an appropriate modification to derive nodes. For example, in sorting algorithms, we treat every input list element as a separate node, and in string matching, we treat each character of the two input strings as a separate node. 

All information over these graphs falls under the following categorisation:

{\bf Stage:} Every feature, i.e. observation in the trajectory, is either part of the \emph{input}, \emph{output}, or the \emph{hints}. As we do not cover algorithms that perform on-line querying, for all 30 algorithms there will be exactly one snapshot of the input and output values, whereas hints will be a time-series of intermediate algorithm states.

{\bf Location:} Every feature is either present within the \emph{nodes}, \emph{edges} (pairs of nodes) or the \emph{graph}\footnote{This also determines shapes of each feature, e.g. node features are of shape $n\times f$; edge features are of shape $n\times n\times f$; graph features are of shape $f$, where $f$ is the dimension of this feature (excluding batch axis).}.

{\bf Type:} Every feature can be of five possible types, which can determine the appropriate method for encoding/decoding it, and the appropriate loss function to use when learning to predict it:
\begin{itemize}
\item {\tt scalar}: Floating-point scalar\footnote{Given our current restriction on numerical predictions, scalar types will never be given in the output stage.} feature. This would typically be fit using mean-squared error.

\item {\tt categorical}: Categorical feature over $K$ possible classes. The type corresponds typically to cross-entropy loss over the classes.

\item {\tt mask}: Categorical feature over two classes. This can be fit using binary cross-entropy.

\item {\tt mask\_one}: Categorical feature over two classes, where exactly one node is active (``one-hot''). One would generally optimise this argmax operation using categorical cross-entropy.

\item {\tt pointer}: Categorical feature over the $n$ nodes. To predict ``similarity'' score against every node, and typically optimised using categorical cross entropy (as introduced in Pointer Graph Networks (PGN)~\cite{velivckovic2020pointer}).
\end{itemize}

Specifying a feature's stage, location and type fully determines its role in the dataflow. A tuple \texttt{(stage, loc, type, values)} is referred to as a \emph{probe}. Each of the 30 algorithms has a static (w.r.t. stage, location and type) set of probes, which are considered to be a \emph{spec} for the algorithm. We will later describe how these specs may be used to construct baseline architectures for the benchmark.

Every node is always endowed with a \emph{position} scalar input probe, which uniquely indexes it---the values are linearly spaced between 0 and 1 along the node index. This allows not only representing the data sequentially (when this is appropriate), but also serves as a useful tie-breaker when algorithms could make an arbitrary choice on which node to explore next---we force the algorithms to favour nodes with \emph{smaller} position values.

To illustrate these concepts further, at the end of this section we will describe the probes in detail for a popular algorithm (insertion sort).

Note that, while we format the data in a way that clearly favours graph neural network executors, it can be easily adapted for different types of neural architectures; for example, sequence to sequence models \citep{sutskever2014sequence}.

Overall, CLRS-30 requires $\sim1\mathrm{h}$ to generate, and occupies $\sim4.5\mathrm{GB}$ when uncompressed, across all 30 tasks.

\subsection{Hints}
\label{sec:hints}

Hints are an important component of our benchmark, which we find fundamental in order to make progress on algorithmic reasoning. As we previously argued, the advantage of algorithms as a task is our understanding of their behaviour, and our ability to decompose them into useful subroutines that can be shared or repeatedly applied. 

While, implicitly, we hope that such a decomposition would happen in any learned system, even when trained just using inputs and outputs (as studied in \citet{xu2019can}), the degree to which we can measure or encourage this is limited in the typical end-to-end learning process, and often most of the generalisation happens only in-distribution (as observed by \citet{velivckovic2019neural,xu2020neural,bevilacqua2021size}). The underlying algorithm may not be statistically identifiable from a small set of input/output pairs.

Conversely, a perfect decomposition of a task into small subtasks can be generated for algorithmic problems. Then, individual models for each subtask may be trained and recomposed into a solution. Such an approach will, by construction, provide strong decompositional benefits: as studied by \citet{yan2020neural}, perfect OOD generalisation can be observed with such models, and they can even generalise zero-shot to test algorithms that reuse their modules. However, the downstream applicability of this is potentially limited; when faced with a novel task which cannot be easily decomposed into subtasks, it can be hard to decide how to reuse the learnt modules.

We believe hints to lie in-between these two approaches. On one hand, they represent intermediate targets which the network should be able to predict if it performs reasoning similar\footnote{Note that architectures supervised in this way usually don't model the hints perfectly, and will deviate from the target algorithm in subtle ways---\citet{velivckovic2020pointer} perform a qualitative study which shows GPU-specialised data structures could emerge as a result of such setups.} to the ground truth algorithm it is supposed to mimic. Indeed, several lines of recent work \citep{velivckovic2019neural,georgiev2020neural,velivckovic2020pointer,deac2020graph} make favourable conclusions about using them, when it comes to achieving stronger OOD generalisation. Furthermore, models leveraging hints are still end-to-end models; when faced with a novel task at test-time, we don't need explicit knowledge of that task's hints in order to re-use the weights learnt on a task which had them.

\begin{figure*}
\includegraphics[width=\linewidth]{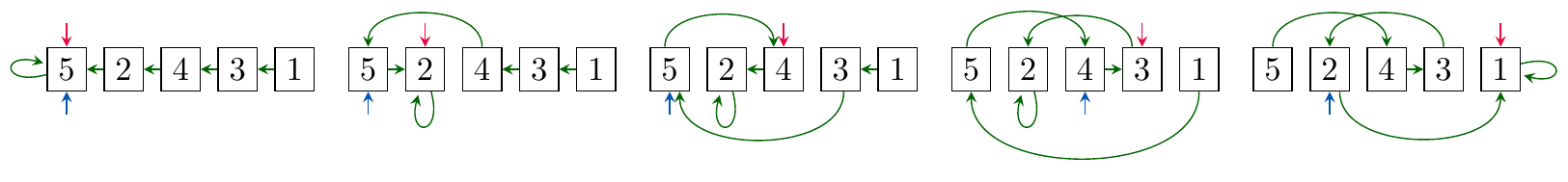}
\caption{A sequence of hints for insertion sorting a list $[5, 2, 4, 3, 1]$. \textbf{\textcolor{olivegreen}{Green}} pointers correspond to the predecessor pointers (specifying the list's state throughout the algorithm's execution. Note how the head of the list always points to itself, by convention. Further, note how, at every step, the list is rewired such that the node selected by the \textbf{\textcolor{echonavy}{blue}} pointer (slot) will point to the current iterator (pointed in \textbf{\textcolor{vividred}{red}}).}

\label{fig:insert}
\end{figure*}

Algorithms specify \emph{one} way of attacking a problem, that is explicitly detailed through the hints. In this sense, insertion sort (to be presented shortly) is one way of implementing a \emph{sorting} function: \textbf{all} sorting algorithms model sorting functions, and will hence have identical outputs for identical inputs. The aspects that set the different sorting algorithms apart are exposed through their hints.

Being mindful of the fact that neural networks commonly run on parallelisable architectures, we have made efforts to ``compress'' the hints as much as possible. For example, if a single \texttt{for} loop is used to sweep the data and detect the node which optimises a certain quantity (without doing any order-sensitive computations), that \texttt{for} loop can typically be entirely ``skipped'' when recording hints: as parallel architectures may typically examine all the nodes at once. Further, we make every effort possible that the hint at step $t+1$ will be predictable from the hints at step $t$ by using only a single step of message passing. 

\subsection{Worked example: insertion sort}

To illustrate all of the concepts outlined above, we observe the trajectories extracted by our data collection procedure on an example: \emph{insertion sorting} the array $[5, 2, 4, 3, 1]$.

Insertion sort uses one pointer ($j$) to scan through the array, and then another pointer ($i$) to slot the $j$-th item into the correct place within $[0..j]$. This ascertains the invariant that, after $k$ steps, the subarray of the first $k$ elements is completely sorted. Hence the trajectory (with $i$ and $j$ marked) is:
$[5_{i,j}, 2, 4, 3, 1] \rightarrow [2_i, 5_j, 4, 3, 1] \rightarrow [2, 4_i, 5_j, 3, 1] \rightarrow [2_, 3_i, 4, 5_j, 1] \rightarrow [1_i, 2, 3, 4, 5_j]$. Here, at each step, $j$ scans along the array, and $i$ indicates the correct place for the element that was $j$-th at the start of each iteration.

Converting this trajectory into a graph representation requires some considerations. Requiring the model to perform explicit swapping of node values would, ultimately, require numerical predictions. To avoid it, we ask the model to predict the \emph{predecessor pointer} of each node (by convention, the head of the array points to itself). Hence the actual recorded trajectory can be realised as depicted in Figure \ref{fig:insert}. In this figure, \textbf{\textcolor{olivegreen}{green}} pointers correspond to the predecessor pointers, \textbf{\textcolor{vividred}{red}} ones point to $j$, and \textbf{\textcolor{echonavy}{blue}} ones point to $i$. $i$ and $j$ are realised as type \texttt{mask\_one}, whereas predecessors are of type \texttt{pointer}---and all three are stored in the nodes. The \textbf{\textcolor{vividred}{red}} and \textbf{\textcolor{echonavy}{blue}} pointers represent the ``hints'' for this task. 

Finally, note that the original insertion sort pseudocode mandates that, at each iteration, $i$ starts at position $j$ and shifts backward until the right position is found. However, this procedure can be performed in one step by a GNN, as it can locate the correct position by examining all relevant positions, and we can omit all of those intermediate steps.

In order to further illustrate how these hints are collected, we also provide an informal pseudocode for collecting hints for insertion sort in Algorithm 1:

\begin{algorithm}[H]\label{algo:xlvin_overview}
\caption{Hint updates for Insertion Sort}
	\SetKwInOut{Input}{Input}\SetKwInOut{Output}{Hints}
    \Input{Input array $\mathtt{val}$, Positions $\mathtt{pos}$}
    \Output{Predecessors $\mathtt{\textcolor{olivegreen}{pred}}$, Iterator $\mathtt{\textcolor{vividred}{iter}}$, swap slot $\mathtt{\textcolor{echonavy}{slot}}$}
    \BlankLine
    $\mathtt{\textcolor{olivegreen}{pred}[i]}\leftarrow\begin{cases}0 & \mathtt{i} = 0\\
    \mathtt{i}-1 & \mathtt{i} > 0
    \end{cases}$  \tcp*{Initialise list}
    \BlankLine
    $\mathtt{\textcolor{echonavy}{slot}}\leftarrow 0, \mathtt{\textcolor{vividred}{iter}}\leftarrow 0$
    \BlankLine
    \While{$\mathtt{\textcolor{vividred}{iter}} < n$}{
          $\mathtt{\textcolor{vividred}{iter}}\leftarrow\mathtt{\textcolor{vividred}{iter}} + 1$\BlankLine
          $\mathtt{max\_node}\leftarrow \operatornamewithlimits{argmax}\limits_{\mathtt{j\ :\ pos[j]<pos[\textcolor{vividred}{iter}]}}\mathtt{val[j]}$\BlankLine
          \uIf{$\mathtt{val[max\_node] < val[\textcolor{vividred}{iter}]}$}{$\mathtt{\textcolor{echonavy}{slot}}\leftarrow\mathtt{max\_node}$\BlankLine$
          \mathtt{\textcolor{olivegreen}{pred}[i]}\leftarrow\begin{cases}\mathtt{\textcolor{echonavy}{slot}} & \mathtt{i} = \mathtt{\textcolor{vividred}{iter}}\\
          \mathtt{\textcolor{olivegreen}{pred}[i]} & \mathrm{otherwise}
          \end{cases}$}
          \Else{
          $\mathtt{\textcolor{echonavy}{slot}}\leftarrow \operatornamewithlimits{argmin}\limits_{\mathtt{j}\ :\ \mathtt{pos[j]}< \mathtt{pos[\textcolor{vividred}{iter}]}, \mathtt{val[j]\geq val[\textcolor{vividred}{iter}]}} \mathtt{val[j]}$\BlankLine
          $\mathtt{\textcolor{olivegreen}{pred}[i]}\leftarrow\begin{cases}
          \mathtt{\textcolor{vividred}{iter}} & \mathtt{i} = \mathtt{\textcolor{echonavy}{slot}} \\
          \mathtt{\textcolor{vividred}{iter}} & {\scriptstyle \mathtt{i=\textcolor{vividred}{iter} \wedge  \textcolor{olivegreen}{pred}[\textcolor{echonavy}{slot}] = \textcolor{echonavy}{slot}}}\\
          \mathtt{\textcolor{olivegreen}{pred}[\textcolor{echonavy}{slot}]} & {\scriptstyle \mathtt{i=\textcolor{vividred}{iter} \wedge \textcolor{olivegreen}{pred}[\textcolor{echonavy}{slot}]\neq\textcolor{echonavy}{slot}}}\\
          \mathtt{max\_node} & \mathtt{\textcolor{olivegreen}{pred}[i] = \textcolor{vividred}{iter}}\\
          \mathtt{\textcolor{olivegreen}{pred}[i]} & \mathrm{otherwise}
          \end{cases}$}
    	}
    	\Return{$\mathtt{\textcolor{olivegreen}{pred}}$} \tcp*{Return final list}
\end{algorithm}

In the interest of illustrating the hint structures further, we provide worked examples of trajectories for three more algorithms (dynamic programming, path-finding and string matching) in Appendix \ref{app:examples}. It should be remarked that we directly expose all of the hint collection routines as Python code inside the CLRS library, allowing for direct inspection.

\section{Empirical evaluation}
\label{sec:experiments}

\begin{figure*}[h]
\includegraphics[width=\textwidth]{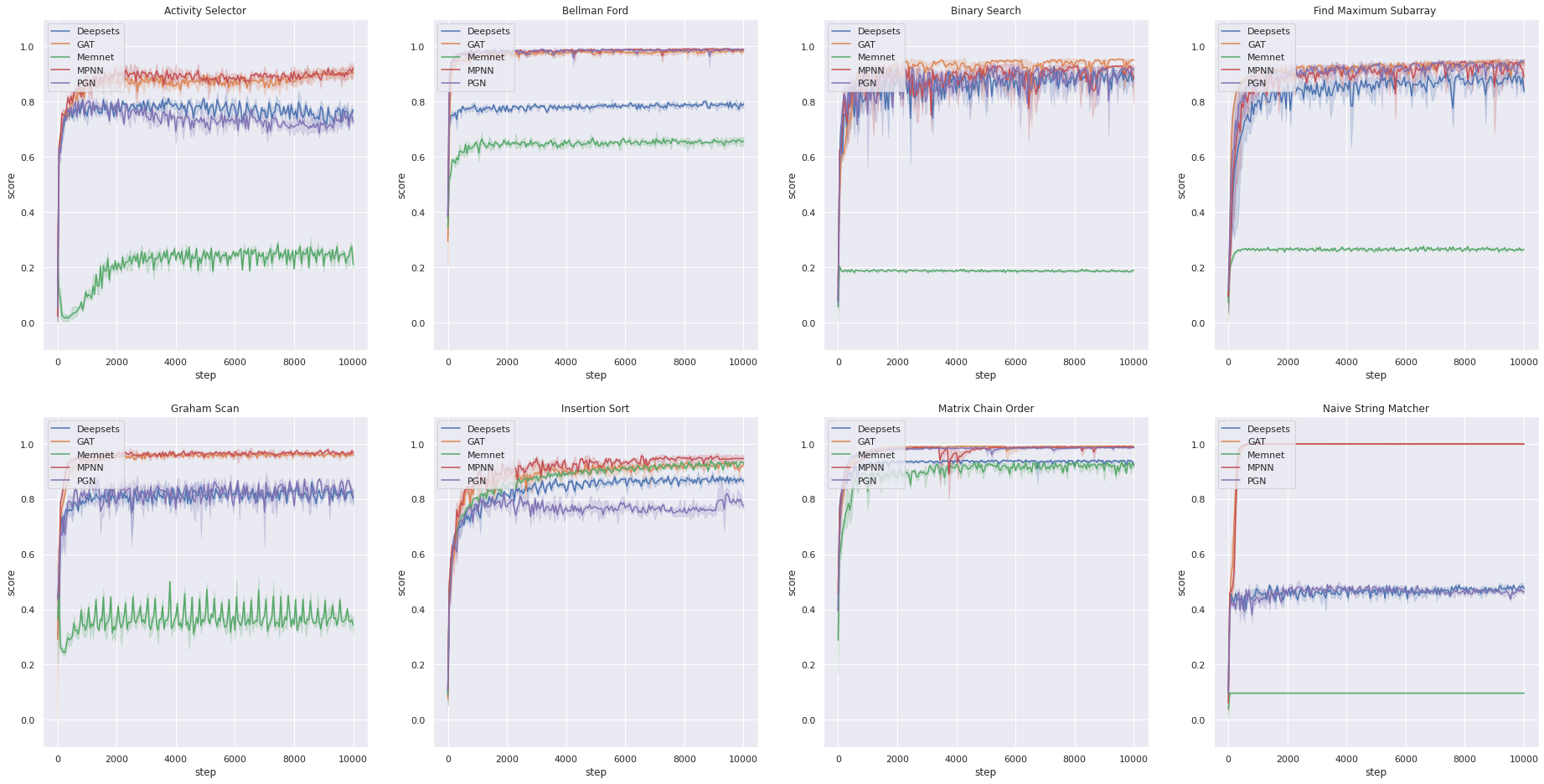}
\caption{Validation results on eight representative algorithms in CLRS-30 (activity selector, Bellman-Ford, binary search, find maximum subarray, Graham scan, insertion sort, matrix chain order, na\"{i}ve string matcher), averaged over three seeds. In all cases the $y$-axis is between $[0,100]\%$. Legend: MPNN \textcolor{vividred}{\bf red}, PGN \textcolor{echopurple}{\bf purple}, Deep Sets \textcolor{echonavy}{\bf blue}, GAT \textcolor{orange}{\bf orange}, Memory Networks \textcolor{olivegreen}{\bf green}. Validation results for all 30 individual algorithms can be found in Appendix \ref{app:vertical}.}
\label{fig:plots}
\end{figure*}

Having surveyed the specifics of CLRS-30, we now present experimental results on it for several proposed algorithmic reasoning models. We primarily investigate whether a natural \emph{ladder} of model performance will emerge when extrapolating to larger inputs. Beyond this, we believe the benchmark will be useful for empirically examining many other properties of algorithmic models, such as evaluating generalisation across different graph types, task types, or various multi-task \citep{xhonneux2021transfer} or continual learning setups. We make available complete implementations of our data generating, probing and model training subroutines, which should make evaluating on such settings simple to deploy\footnote{\url{https://github.com/deepmind/clrs}}. We survey several key ways of interacting with the benchmark (e.g. implementing baselines, modifying datasets, adding new algorithms) in Appendix \ref{app:interface}.

\subsection{Baseline models}

\paragraph{Encode-process-decode} For our experimental validation, we adopt the encode-process-decode paradigm of \citet{hamrick2018relational}, which is a common direction for several hint-based architectures \citep{velivckovic2019neural,georgiev2020neural,velivckovic2020pointer,deac2020graph}.

Namely, we consider a setup with inputs $\vec{x}_i$ in nodes, $\vec{e}_{ij}$ in edges, and $\vec{g}$ in the graph. We first encode each of these using linear layers $f_n, f_e, f_g$, to obtain encodings 
\begin{equation}
\vec{h}_i = f_n(\vec{x}_i) \qquad \vec{h}_{ij} = f_e(\vec{e}_{ij}) \qquad \vec{h}_g = f_g(\vec{g})
\end{equation}
We then feed these latents through a \emph{processor network} to perform one step of computation. As we are focusing on graph representation learning in the current data format, most of our processors will be realised as graph neural networks \citep{gilmer2017neural}. Most generally, along every edge $(i, j)$, a \emph{message} from node $i$ to node $j$, $\vec{m}_{ij}$ is computed (using a message function $f_m$), and these messages are then aggregated across all neighbouring nodes using a permutation-invariant aggregation function, $\bigoplus$. Finally, a \emph{readout network} $f_r$ transforms these aggregated messages and the node encodings into processed node encodings:
\begin{align}\label{eqn:mpn}
    \vec{m}_{ij} &= f_m(\vec{h}_i, \vec{h}_j, \vec{h}_{ij}, \vec{h}_g)\\ \vec{m}_i &= \bigoplus_{i\in\mathcal{N}_j} \vec{m}_{ji} \qquad \qquad  \vec{h}'_i = f_r(\vec{h}_i, \vec{m}_i)
\end{align}
Once node encodings are updated, we can \emph{decode} them to make various predictions for this step of reasoning, depending on the type of the prediction required (using relevant decoder functions $g_\cdot$), as prescribed in Section \ref{sect:impl}. Further, we keep track of previous-step node encodings $\vec{h}^{(t-1)}_i$, to explicitly use in a recurrent cell update (exactly as done by \citet{velivckovic2019neural}). We opt to provide this recurrent update in order to provide long-range capacity to the model.

Lastly, we need to decide in what capacity will hints be used. We provide results for the option where hints are both decoded (used for computing the loss function) and encoded (considered as part of $\vec{x}$, $\vec{e}_{ij}$ and $\vec{g}$). At testing time, the encoded hint is equal to the hints decoded by the previous step, whereas we can stabilise these trajectories at training time by performing \emph{noisy teacher forcing}---inspired by Noisy Nodes \citep{godwin2021simple}, at each step we feed back ground-truth hints with probability $0.5$.  The quantity of hints is still used to determine the number of processor steps to perform at evaluation time. This requirement of knowing the hint-size can be lifted by, e.g., using termination networks \citep{velivckovic2019neural,banino2021pondernet} or aligning to iterative algorithms \citep{tang2020towards}. 

\begin{table*}[ht]
    \centering

    \caption{Average test micro-F$_1$ score of all models on all algorithm classes. The full test results for all 30 algorithms, along with a breakdown of the ``win/tie/loss'' metric, are given in Appendix \ref{app:testfull}.}
    \begin{tabular}{lccccc} \toprule
    \textbf{Algorithm} & \textbf{Deep Sets} & \textbf{GAT} & \textbf{Memnet} & \textbf{MPNN} & \textbf{PGN} \\\midrule
Divide \& Conquer & $ 12.48\% \pm  0.67$ & $ 24.43\% \pm  0.74$ & $ 13.05\% \pm  0.00$ & $ 20.30\% \pm  0.85$ & $\mathbf{ 65.23\%} \pm  4.44$\\
Dynamic Prog. & $ 66.05\% \pm  7.79$ & $ 67.19\% \pm  5.33$ & $ 67.94\% \pm  7.75$ & $ 65.10\% \pm  6.44$ & $\mathbf{ 70.58\%} \pm  6.48$\\
Geometry & $ 64.08\% \pm  6.60$ & $\mathbf{ 73.27\%} \pm  11.18$ & $ 45.14\% \pm  11.65$ & $ 73.11\% \pm  17.19$ & $ 61.19\% \pm  7.01$\\
Graphs & $ 37.65\% \pm  8.09$ & $ 46.80\% \pm  8.66$ & $ 24.12\% \pm  5.20$ & $\mathbf{ 62.79\%} \pm  8.75$ & $ 60.25\% \pm  8.42$\\
Greedy & $ 75.47\% \pm  6.81$ & $ 78.96\% \pm  4.59$ & $ 53.42\% \pm  20.73$ & $\mathbf{ 82.39\%} \pm  3.01$ & $ 75.84\% \pm  6.59$\\
Search & $ 43.79\% \pm  18.29$ & $ 37.35\% \pm  19.81$ & $ 34.35\% \pm  21.67$ & $ 41.20\% \pm  19.87$ & $\mathbf{ 56.11\%} \pm  21.56$\\
Sorting & $ 39.60\% \pm  7.19$ & $ 14.35\% \pm  4.64$ & $\mathbf{ 71.53\%} \pm  1.09$ & $ 11.83\% \pm  2.78$ & $ 15.45\% \pm  8.46$\\
Strings & $ 2.64\% \pm  0.68$ & $ 3.02\% \pm  1.08$ & $ 1.51\% \pm  0.21$ & $\mathbf{ 3.21\%} \pm  0.94$ & $ 2.04\% \pm  0.20$\\
\midrule
Overall average  & $ 42.72\%$ & $ 43.17\%$ & $ 38.88\%$ & $ 44.99\%$ & $\mathbf{ 50.84\%}$\\
Win/Tie/Loss counts & $0/3/27$ & $1/5/24$ & $4/2/24$ & $8/3/19$ & $\bf 8/6/16$\\ \bottomrule
\end{tabular}
   
    \label{tab:test_results_all}
    \end{table*}

\paragraph{Processor networks} The only remaining component to specify is the \emph{processor network} used by our models. As this component carries the most computational load, it is also the most obvious module to sweep over. We provide all implementations and hyperparameters within our codebase.

Unless otherwise specified, we assume \emph{fully-connected graphs}, i.e. $\mathcal{N}_i = \{1, 2, \dots, n\}$, hence every node is connected to every other node. We consider the following baseline processor networks:

{\bf Deep Sets} \citep{zaheer2017deep}; where each node is only connected to itself: $\mathcal{N}_i = \{i\}$ (i.e., choice of $\bigoplus$ is irrelevant). Such a model is popular for summary statistic tasks.

{\bf Graph Attention Networks} \citep{velivckovic2017graph}, where the aggregation function $\bigoplus$ is self-attention \citep{vaswani2017attention}, and the message function $f_m$ merely extracts the sender features: $f_m(\vec{h}_i, \vec{h}_j, \vec{h}_{ij}, \vec{h}_g) = {\bf W}\vec{h}_i$. We report the best performance across GAT \citep{velivckovic2017graph} and GATv2 \citep{brody2021attentive} attention mechanisms.\

{\bf Message-passing Neural Networks} \citep{gilmer2017neural}, which correspond exactly to the formulation in Equation \ref{eqn:mpn}, with $\bigoplus=\max$, as prescribed by previous work \citep{velivckovic2019neural}. As a sanity check, we also attempted $\bigoplus=\sum$ finding it underperformed on all tasks compared to $\max$.

{\bf Pointer Graph Networks} \citep{velivckovic2020pointer}, which use only graph neighbourhoods $\mathcal{N}_i$ specified by a union of all node \texttt{pointer} and edge \texttt{mask} hints, and $\bigoplus=\max$. This restricts the model to only reason over the edges deemed important by the inputs and hints.

{\bf Memory Networks} \citep{sukhbaatar2015end} have been used in the past as baseline for investigating reasoning in neural networks \citep[e.g.][]{banino2020memo}, as they provide an alternative way to use structural dependencies in a graph by treating edges as memories and nodes as queries. Here we used latents representing node features $\vec{h}_i$ as queries and latents representing edge features $\vec{h}_{ij}$ (where there is a connecting edge and $\vec{0}$ otherwise) as memory inputs.

\subsection{Dataset statistics}

For each algorithm in CLRS-30, we provide a canonical set of training, validation and test trajectories for benchmarking in- and out-of-distribution generalisation. We obtain these trajectories by running the algorithms on randomly sampled inputs that conform to their input specification. This implies, e.g., that the inputs to most graph algorithms are Erd\H{o}s-R\'{e}nyi graphs \citep{erdos2011evolution} with a certain edge probability. All scalar inputs are sampled from $U(0, 1)$. 

For validation, our aim is to measure in-distribution generalisation. Hence we sample inputs of 16 nodes for both, and generate 1,000 trajectories for training and 32 for validation. For testing, we measure out-of-distribution generalisation, and sample 32 trajectories for inputs of 64 nodes. For algorithms where the output is on the graph stage (rather than node/edge), we generate $64\times$ more trajectories, in order to equalise the number of targets across tasks.

We optimise our models on the training trajectories in a teacher-forced fashion, with a batch size of 32, using the Adam optimiser \citep{kingma2014adam} with an initial learning rate of $\eta=0.001$. We train for $10,000$ steps, early stopping on the validation performance. Our models are trained on one V100 Volta GPU, requiring roughly between 1h and 30h to train, depending on the algorithm's time complexity. For example, linear-time algorithms have significantly fewer hints---hence message passing steps---than cubic-time ones.

\subsection{Validation (in-distribution) performance}

We provide the in-distribution performance throughout training in Figure \ref{fig:plots}, for eight representative tasks in CLRS-30 (one per each algorithm type); see Appendix \ref{app:vertical} for the full results on all 30 algorithms. In this regime, the MPNN appears to dominate for most tasks: achieving over $90\%$ F$_1$ score for nearly all of them.

While this might seem like strong evidence in favour of the fully-connected MPNNs, their added degrees of freedom may also make MPNNs more prone to overfitting to specifics of the input (e.g. the input graphs' sizes), rather than truly learning the underlying reasoning rule. We present the out-of-distribution results next, in order to make this distinction clear.

\subsection{Test (out-of-distribution) performance}

The averaged out-of-distribution performance (using the early-stopped model on validation) across each of the eight algorithm types is provided in Table \ref{tab:test_results_all}; see Appendix \ref{app:testfull} for the full results on all 30 algorithms. MPNNs are unable to transfer their impressive gains to graphs that are four times larger: in fact, the PGN takes over as the most performant model when averaged across task types---this aligns well with prior research \citep{velivckovic2020pointer}. The outperformance is also observed when we count how frequently each model is among the best-performing models for a given algorithm, as per our ``win/tie/loss'' metric, which we explain in Appendix \ref{app:testfull}. GNN models, additionally, outperform models like Deep Sets and Memory Nets, reinforcing that GNNs are a useful primitive for algorithmic reasoning \citep{xu2019can,dudzik2022graph}.

Aside from all of the above, we note that the OOD version of the CLRS-30 benchmark is highly challenging and far from solved for most tasks, making it a meaningful informant of future progress in the area. In particular, PGNs struggled on tasks requiring long-range rollouts (such as DFS), or recursive reasoning (such as Quicksort and Quickselect). This invites further research in algorithmic reasoners that can support such computation. It is further revealed that more specialised inductive biases and training regimes may be required to deal with string matching algorithms (such as KMP), and that the processor studied here tended to perform the best on tasks which were of favourable (sublinear) complexity in terms of hint counts (such as BFS, Bellman-Ford, and task scheduling).

The specific results we obtain with our baselines validate several bits of prior research in the area, but also demonstrate we still have a long way to go, with even simple OOD scenarios only being fit to about 50\% micro-F$_1$ performance.

\section{Conclusion}
We introduce CLRS-30, a dataset that contains trajectories from 30 classical algorithms.  This benchmark constitutes an effective way to test out-of-distribution generalization and transfer, and brings a means to evaluate algorithmic reasoning learnt by neural network models. The dataset provides input/output pairs for all algorithms, as well as intermediate trajectory information (``hints'').

It is our hope that CLRS-30 will be a useful tool to shepherd future research in algorithmic reasoning, as prior art in the area largely generated their own datasets, making progress tracking challenging. Further, we hope that CLRS-30 will make algorithmic reasoning a more accessible area: one does not need a background in theoretical computer science to generate the dataset, and can focus on the modelling.

If we convinced you to try out our library, please consult Appendix \ref{app:interface} for detailed instructions on most common ways to interact with our platform. CLRS is in constant development, and we welcome any and all feedback.

\section*{Acknowledgements}
CLRS-30 was developed over a long time-frame, with many useful contributions, which we kindly acknowledge here.

We would like to particularly thank Borja Ibarz for numerous fixes and additions, and laying foundation for future iterations. Additionally, we warmly thank Jonathan Godwin, Sadegh Mahdavi, Euan Ong, MohamedElfatih Salah, Ahmed Elhag, Andreea Deac, Frederik Nijweide, Andrew Dudzik, Thomas Kipf, Amin Barekatain and Dobrik Georgiev for their support, and identifying numerous bugs during development. Finally, we thank Kim Stachenfeld, Nate Kushman and Daan Wierstra for reviewing the paper prior to submission, and anonymous reviewers for their careful feedback, strengthening the paper significantly.

\bibliographystyle{icml2022}
\bibliography{iclr2021_conference}

\begin{thebibliography}{78}
\providecommand{\natexlab}[1]{#1}
\providecommand{\url}[1]{\texttt{#1}}
\expandafter\ifx\csname urlstyle\endcsname\relax
  \providecommand{\doi}[1]{doi: #1}\else
  \providecommand{\doi}{doi: \begingroup \urlstyle{rm}\Url}\fi

\bibitem[Aho et~al.(1974)Aho, Hopcroft, and Ullman]{aho1974design}
Aho, A.~V., Hopcroft, J.~E., and Ullman, J.~D.
\newblock The design and analysis of computer algorithms.
\newblock \emph{Reading}, 1974.

\bibitem[Alet et~al.(2018)Alet, Lozano-Perez, and Kaelbling]{alet18a}
Alet, F., Lozano-Perez, T., and Kaelbling, L.~P.
\newblock Modular meta-learning.
\newblock volume~87 of \emph{Proceedings of Machine Learning Research}. PMLR,
  2018.

\bibitem[Banino et~al.(2020)Banino, Badia, Köster, Chadwick, Zambaldi,
  Hassabis, Barry, Botvinick, Kumaran, and Blundell]{banino2020memo}
Banino, A., Badia, A.~P., Köster, R., Chadwick, M.~J., Zambaldi, V., Hassabis,
  D., Barry, C., Botvinick, M., Kumaran, D., and Blundell, C.
\newblock Memo: A deep network for flexible combination of episodic memories.
\newblock In \emph{International Conference on Learning Representations}, 2020.
\newblock URL \url{https://openreview.net/forum?id=rJxlc0EtDr}.

\bibitem[Banino et~al.(2021)Banino, Balaguer, and
  Blundell]{banino2021pondernet}
Banino, A., Balaguer, J., and Blundell, C.
\newblock Pondernet: Learning to ponder.
\newblock \emph{arXiv preprint arXiv:2107.05407}, 2021.

\bibitem[Belkin et~al.(2019)Belkin, Hsu, and Xu]{belkin2019two}
Belkin, M., Hsu, D., and Xu, J.
\newblock Two models of double descent for weak features.
\newblock \emph{arXiv preprint arXiv:1903.07571}, 2019.

\bibitem[Bellemare et~al.(2013)Bellemare, Naddaf, Veness, and
  Bowling]{bellemare2013arcade}
Bellemare, M.~G., Naddaf, Y., Veness, J., and Bowling, M.
\newblock The arcade learning environment: An evaluation platform for general
  agents.
\newblock \emph{Journal of Artificial Intelligence Research}, 47:\penalty0
  253--279, 2013.

\bibitem[Bellman(1958)]{bellman1958routing}
Bellman, R.
\newblock On a routing problem.
\newblock \emph{Quarterly of applied mathematics}, 16\penalty0 (1):\penalty0
  87--90, 1958.

\bibitem[Bengio et~al.(2020)Bengio, Lodi, and Prouvost]{bengio2020machine}
Bengio, Y., Lodi, A., and Prouvost, A.
\newblock Machine learning for combinatorial optimization: a methodological
  tour d’horizon.
\newblock \emph{European Journal of Operational Research}, 2020.

\bibitem[Bentley(1984)]{bentley1984programming}
Bentley, J.
\newblock Programming pearls: algorithm design techniques.
\newblock \emph{Communications of the ACM}, 27\penalty0 (9):\penalty0 865--873,
  1984.

\bibitem[Bevilacqua et~al.(2021)Bevilacqua, Zhou, and
  Ribeiro]{bevilacqua2021size}
Bevilacqua, B., Zhou, Y., and Ribeiro, B.
\newblock Size-invariant graph representations for graph classification
  extrapolations.
\newblock In \emph{International Conference on Machine Learning}, pp.\
  837--851. PMLR, 2021.

\bibitem[Brody et~al.(2021)Brody, Alon, and Yahav]{brody2021attentive}
Brody, S., Alon, U., and Yahav, E.
\newblock How attentive are graph attention networks?
\newblock \emph{arXiv preprint arXiv:2105.14491}, 2021.

\bibitem[Cappart et~al.(2021)Cappart, Ch{\'e}telat, Khalil, Lodi, Morris, and
  Veli{\v{c}}kovi{\'c}]{cappart2021combinatorial}
Cappart, Q., Ch{\'e}telat, D., Khalil, E., Lodi, A., Morris, C., and
  Veli{\v{c}}kovi{\'c}, P.
\newblock Combinatorial optimization and reasoning with graph neural networks.
\newblock \emph{arXiv preprint arXiv:2102.09544}, 2021.

\bibitem[Chen et~al.(2020)Chen, Chen, Villar, and Bruna]{chen2020can}
Chen, Z., Chen, L., Villar, S., and Bruna, J.
\newblock Can graph neural networks count substructures?
\newblock \emph{arXiv preprint arXiv:2002.04025}, 2020.

\bibitem[Cormen et~al.(2009)Cormen, Leiserson, Rivest, and
  Stein]{cormen2009introduction}
Cormen, T.~H., Leiserson, C.~E., Rivest, R.~L., and Stein, C.
\newblock \emph{Introduction to algorithms}.
\newblock MIT press, 2009.

\bibitem[Corso et~al.(2020)Corso, Cavalleri, Beaini, Li{\`o}, and
  Veli{\v{c}}kovi{\'c}]{corso2020principal}
Corso, G., Cavalleri, L., Beaini, D., Li{\`o}, P., and Veli{\v{c}}kovi{\'c}, P.
\newblock Principal neighbourhood aggregation for graph nets.
\newblock \emph{arXiv preprint arXiv:2004.05718}, 2020.

\bibitem[Deac et~al.(2020)Deac, Bacon, and Tang]{deac2020graph}
Deac, A., Bacon, P.-L., and Tang, J.
\newblock Graph neural induction of value iteration.
\newblock \emph{arXiv preprint arXiv:2009.12604}, 2020.

\bibitem[Deac et~al.(2021)Deac, Veli{\v{c}}kovi{\'c}, Milinkovic, Bacon, Tang,
  and Nikolic]{deac2020xlvin}
Deac, A.-I., Veli{\v{c}}kovi{\'c}, P., Milinkovic, O., Bacon, P.-L., Tang, J.,
  and Nikolic, M.
\newblock Neural algorithmic reasoners are implicit planners.
\newblock \emph{Advances in Neural Information Processing Systems}, 34, 2021.

\bibitem[Dijkstra et~al.(1959)]{dijkstra1959note}
Dijkstra, E.~W. et~al.
\newblock A note on two problems in connexion with graphs.
\newblock \emph{Numerische mathematik}, 1\penalty0 (1):\penalty0 269--271,
  1959.

\bibitem[Du et~al.(2018)Du, Czarnecki, Jayakumar, Pascanu, and
  Lakshminarayanan]{du2018adapting}
Du, Y., Czarnecki, W.~M., Jayakumar, S.~M., Pascanu, R., and Lakshminarayanan,
  B.
\newblock Adapting auxiliary losses using gradient similarity, 2018.

\bibitem[Dudzik \& Veli{\v{c}}kovi{\'c}(2022)Dudzik and
  Veli{\v{c}}kovi{\'c}]{dudzik2022graph}
Dudzik, A. and Veli{\v{c}}kovi{\'c}, P.
\newblock Graph neural networks are dynamic programmers.
\newblock \emph{arXiv preprint arXiv:2203.15544}, 2022.

\bibitem[Dwivedi et~al.(2020)Dwivedi, Joshi, Laurent, Bengio, and
  Bresson]{dwivedi2020benchmarking}
Dwivedi, V.~P., Joshi, C.~K., Laurent, T., Bengio, Y., and Bresson, X.
\newblock Benchmarking graph neural networks.
\newblock \emph{arXiv preprint arXiv:2003.00982}, 2020.

\bibitem[Erd{\"o}s \& R{\'e}nyi(2011)Erd{\"o}s and
  R{\'e}nyi]{erdos2011evolution}
Erd{\"o}s, P. and R{\'e}nyi, A.
\newblock On the evolution of random graphs.
\newblock In \emph{The structure and dynamics of networks}, pp.\  38--82.
  Princeton University Press, 2011.

\bibitem[Floyd(1962)]{floyd1962algorithm}
Floyd, R.~W.
\newblock Algorithm 97: shortest path.
\newblock \emph{Communications of the ACM}, 5\penalty0 (6):\penalty0 345, 1962.

\bibitem[Ford~Jr \& Fulkerson(2015)Ford~Jr and Fulkerson]{ford2015flows}
Ford~Jr, L.~R. and Fulkerson, D.~R.
\newblock \emph{Flows in networks}.
\newblock Princeton university press, 2015.

\bibitem[Freivalds et~al.(2019)Freivalds, Ozoli{\c{n}}{\v{s}}, and
  {\v{S}}ostaks]{freivalds2019neural}
Freivalds, K., Ozoli{\c{n}}{\v{s}}, E., and {\v{S}}ostaks, A.
\newblock Neural shuffle-exchange networks-sequence processing in o (n log n)
  time.
\newblock In \emph{Advances in Neural Information Processing Systems}, pp.\
  6630--6641, 2019.

\bibitem[Gavril(1972)]{gavril1972algorithms}
Gavril, F.
\newblock Algorithms for minimum coloring, maximum clique, minimum covering by
  cliques, and maximum independent set of a chordal graph.
\newblock \emph{SIAM Journal on Computing}, 1\penalty0 (2):\penalty0 180--187,
  1972.

\bibitem[Georgiev \& Li\'{o}(2020)Georgiev and Li\'{o}]{georgiev2020neural}
Georgiev, D. and Li\'{o}, P.
\newblock Neural bipartite matching.
\newblock \emph{arXiv preprint arXiv:2005.11304}, 2020.

\bibitem[Gilmer et~al.(2017)Gilmer, Schoenholz, Riley, Vinyals, and
  Dahl]{gilmer2017neural}
Gilmer, J., Schoenholz, S.~S., Riley, P.~F., Vinyals, O., and Dahl, G.~E.
\newblock Neural message passing for quantum chemistry.
\newblock \emph{arXiv preprint arXiv:1704.01212}, 2017.

\bibitem[Godwin et~al.(2021)Godwin, Schaarschmidt, Gaunt, Sanchez-Gonzalez,
  Rubanova, Veli{\v{c}}kovi{\'c}, Kirkpatrick, and Battaglia]{godwin2021simple}
Godwin, J., Schaarschmidt, M., Gaunt, A.~L., Sanchez-Gonzalez, A., Rubanova,
  Y., Veli{\v{c}}kovi{\'c}, P., Kirkpatrick, J., and Battaglia, P.
\newblock Simple gnn regularisation for 3d molecular property prediction and
  beyond.
\newblock In \emph{International Conference on Learning Representations}, 2021.

\bibitem[Graham(1972)]{graham1972efficient}
Graham, R.~L.
\newblock An efficient algorithm for determining the convex hull of a finite
  planar set.
\newblock \emph{Info. Pro. Lett.}, 1:\penalty0 132--133, 1972.

\bibitem[Graves et~al.(2014)Graves, Wayne, and Danihelka]{graves2014neural}
Graves, A., Wayne, G., and Danihelka, I.
\newblock Neural turing machines.
\newblock \emph{arXiv preprint arXiv:1410.5401}, 2014.

\bibitem[Griffiths et~al.(2019)Griffiths, Callaway, Chang, Grant, Krueger, and
  Lieder]{Griffiths19}
Griffiths, T., Callaway, F., Chang, M., Grant, E., Krueger, P., and Lieder, F.
\newblock Doing more with less: meta-reasoning and meta-learning in humans and
  machines.
\newblock \emph{Current Opinion in Behavioral Sciences}, October 2019.

\bibitem[Hamrick et~al.(2018)Hamrick, Allen, Bapst, Zhu, McKee, Tenenbaum, and
  Battaglia]{hamrick2018relational}
Hamrick, J.~B., Allen, K.~R., Bapst, V., Zhu, T., McKee, K.~R., Tenenbaum,
  J.~B., and Battaglia, P.~W.
\newblock Relational inductive bias for physical construction in humans and
  machines.
\newblock \emph{arXiv preprint arXiv:1806.01203}, 2018.

\bibitem[Hennigan et~al.(2020)Hennigan, Cai, Norman, and
  Babuschkin]{haiku2020github}
Hennigan, T., Cai, T., Norman, T., and Babuschkin, I.
\newblock {H}aiku: {S}onnet for {JAX}, 2020.
\newblock URL \url{http://github.com/deepmind/dm-haiku}.

\bibitem[Hierholzer \& Wiener(1873)Hierholzer and
  Wiener]{hierholzer1873moglichkeit}
Hierholzer, C. and Wiener, C.
\newblock {\"U}ber die m{\"o}glichkeit, einen linienzug ohne wiederholung und
  ohne unterbrechung zu umfahren.
\newblock \emph{Mathematische Annalen}, 6\penalty0 (1):\penalty0 30--32, 1873.

\bibitem[Hoare(1961)]{hoare1961algorithm}
Hoare, C.~A.
\newblock Algorithm 65: find.
\newblock \emph{Communications of the ACM}, 4\penalty0 (7):\penalty0 321--322,
  1961.

\bibitem[Hoare(1962)]{hoare1962quicksort}
Hoare, C.~A.
\newblock Quicksort.
\newblock \emph{The Computer Journal}, 5\penalty0 (1):\penalty0 10--16, 1962.

\bibitem[Jacot et~al.(2018)Jacot, Gabriel, and Hongler]{Jacot2018NTK}
Jacot, A., Gabriel, F., and Hongler, C.
\newblock Neural tangent kernel: Convergence and generalization in neural
  networks.
\newblock In \emph{Advances in Neural Information Processing Systems 31}. 2018.

\bibitem[Jarvis(1973)]{jarvis1973identification}
Jarvis, R.~A.
\newblock On the identification of the convex hull of a finite set of points in
  the plane.
\newblock \emph{Information processing letters}, 2\penalty0 (1):\penalty0
  18--21, 1973.

\bibitem[Joshi et~al.(2020)Joshi, Cappart, Rousseau, Laurent, and
  Bresson]{joshi2020learning}
Joshi, C.~K., Cappart, Q., Rousseau, L.-M., Laurent, T., and Bresson, X.
\newblock Learning tsp requires rethinking generalization.
\newblock \emph{arXiv preprint arXiv:2006.07054}, 2020.

\bibitem[Kaiser \& Sutskever(2015)Kaiser and Sutskever]{kaiser2015neural}
Kaiser, {\L}. and Sutskever, I.
\newblock Neural gpus learn algorithms.
\newblock \emph{arXiv preprint arXiv:1511.08228}, 2015.

\bibitem[Khalil et~al.(2017)Khalil, Dai, Zhang, Dilkina, and
  Song]{khalil2017learning}
Khalil, E., Dai, H., Zhang, Y., Dilkina, B., and Song, L.
\newblock Learning combinatorial optimization algorithms over graphs.
\newblock In \emph{Advances in Neural Information Processing Systems}, pp.\
  6348--6358, 2017.

\bibitem[Kingma \& Ba(2014)Kingma and Ba]{kingma2014adam}
Kingma, D.~P. and Ba, J.
\newblock Adam: A method for stochastic optimization.
\newblock \emph{arXiv preprint arXiv:1412.6980}, 2014.

\bibitem[Knuth(1973)]{knuth1973fundamental}
Knuth, D.~E.
\newblock Fundamental algorithms.
\newblock 1973.

\bibitem[Knuth et~al.(1977)Knuth, Morris, and Pratt]{knuth1977fast}
Knuth, D.~E., Morris, Jr, J.~H., and Pratt, V.~R.
\newblock Fast pattern matching in strings.
\newblock \emph{SIAM journal on computing}, 6\penalty0 (2):\penalty0 323--350,
  1977.

\bibitem[Kool et~al.(2018)Kool, van Hoof, and Welling]{kool2018attention}
Kool, W., van Hoof, H., and Welling, M.
\newblock Attention, learn to solve routing problems!
\newblock \emph{arXiv preprint arXiv:1803.08475}, 2018.

\bibitem[Kruskal(1956)]{kruskal1956shortest}
Kruskal, J.~B.
\newblock On the shortest spanning subtree of a graph and the traveling
  salesman problem.
\newblock \emph{Proceedings of the American Mathematical society}, 7\penalty0
  (1):\penalty0 48--50, 1956.

\bibitem[Lake(2019)]{Lake19}
Lake, B.~M.
\newblock Compositional generalization through meta sequence-to-sequence
  learning.
\newblock In \emph{Advances in Neural Information Processing Systems 32}, pp.\
  9791--9801. 2019.

\bibitem[Lawler(1985)]{lawler1985traveling}
Lawler, E.~L.
\newblock The traveling salesman problem: a guided tour of combinatorial
  optimization.
\newblock \emph{Wiley-Interscience Series in Discrete Mathematics}, 1985.

\bibitem[Li et~al.(2020)Li, Gimeno, Kohli, and Vinyals]{li2020strong}
Li, Y., Gimeno, F., Kohli, P., and Vinyals, O.
\newblock Strong generalization and efficiency in neural programs.
\newblock \emph{arXiv preprint arXiv:2007.03629}, 2020.

\bibitem[Liu et~al.(2020)Liu, Zhu, and Belkin]{Liu2020TheoryDL}
Liu, C., Zhu, L., and Belkin, M.
\newblock Toward a theory of optimization for over-parameterized systems of
  non-linear equations: the lessons of deep learning.
\newblock \emph{CoRR}, abs/2003.00307, 2020.

\bibitem[Merity et~al.(2016)Merity, Xiong, Bradbury, and
  Socher]{merity2016pointer}
Merity, S., Xiong, C., Bradbury, J., and Socher, R.
\newblock Pointer sentinel mixture models.
\newblock \emph{arXiv preprint arXiv:1609.07843}, 2016.

\bibitem[Mikolov et~al.(2011)Mikolov, Deoras, Kombrink, Burget, and
  Cernocký]{Mikolov11}
Mikolov, T., Deoras, A., Kombrink, S., Burget, L., and Cernocký, J.
\newblock Empirical evaluation and combination of advanced language modeling
  techniques.
\newblock In \emph{INTERSPEECH}, pp.\  605--608, 2011.

\bibitem[Moore(1959)]{moore1959shortest}
Moore, E.~F.
\newblock The shortest path through a maze.
\newblock In \emph{Proc. Int. Symp. Switching Theory, 1959}, pp.\  285--292,
  1959.

\bibitem[Prim(1957)]{prim1957shortest}
Prim, R.~C.
\newblock Shortest connection networks and some generalizations.
\newblock \emph{The Bell System Technical Journal}, 36\penalty0 (6):\penalty0
  1389--1401, 1957.

\bibitem[Richter \& Wattenhofer(2020)Richter and
  Wattenhofer]{richter2020normalized}
Richter, O. and Wattenhofer, R.
\newblock Normalized attention without probability cage.
\newblock \emph{arXiv preprint arXiv:2005.09561}, 2020.

\bibitem[Russakovsky et~al.(2015)Russakovsky, Deng, Su, Krause, Satheesh, Ma,
  Huang, Karpathy, Khosla, Bernstein, Berg, and Fei-Fei]{ILSVRC15}
Russakovsky, O., Deng, J., Su, H., Krause, J., Satheesh, S., Ma, S., Huang, Z.,
  Karpathy, A., Khosla, A., Bernstein, M., Berg, A.~C., and Fei-Fei, L.
\newblock {ImageNet Large Scale Visual Recognition Challenge}.
\newblock \emph{International Journal of Computer Vision (IJCV)}, 115\penalty0
  (3):\penalty0 211--252, 2015.
\newblock \doi{10.1007/s11263-015-0816-y}.

\bibitem[Strathmann et~al.(2021)Strathmann, Barekatain, Blundell, and
  Veli{\v{c}}kovi{\'c}]{strathmann2021persistent}
Strathmann, H., Barekatain, M., Blundell, C., and Veli{\v{c}}kovi{\'c}, P.
\newblock Persistent message passing.
\newblock \emph{arXiv preprint arXiv:2103.01043}, 2021.

\bibitem[Sukhbaatar et~al.(2015)Sukhbaatar, Szlam, Weston, and
  Fergus]{sukhbaatar2015end}
Sukhbaatar, S., Szlam, A., Weston, J., and Fergus, R.
\newblock End-to-end memory networks.
\newblock \emph{arXiv preprint arXiv:1503.08895}, 2015.

\bibitem[Sutskever et~al.(2014)Sutskever, Vinyals, and
  Le]{sutskever2014sequence}
Sutskever, I., Vinyals, O., and Le, Q.~V.
\newblock Sequence to sequence learning with neural networks.
\newblock In \emph{Advances in neural information processing systems}, pp.\
  3104--3112, 2014.

\bibitem[Tang et~al.(2020)Tang, Huang, Gu, Lu, and Su]{tang2020towards}
Tang, H., Huang, Z., Gu, J., Lu, B., and Su, H.
\newblock Towards scale-invariant graph-related problem solving by iterative
  homogeneous gnns.
\newblock \emph{the 34th Annual Conference on Neural Information Processing
  Systems (NeurIPS)}, 2020.

\bibitem[Trask et~al.(2018)Trask, Hill, Reed, Rae, Dyer, and
  Blunsom]{trask2018neural}
Trask, A., Hill, F., Reed, S.~E., Rae, J., Dyer, C., and Blunsom, P.
\newblock Neural arithmetic logic units.
\newblock In \emph{Advances in Neural Information Processing Systems}, pp.\
  8035--8044, 2018.

\bibitem[Vaswani et~al.(2017)Vaswani, Shazeer, Parmar, Uszkoreit, Jones, Gomez,
  Kaiser, and Polosukhin]{vaswani2017attention}
Vaswani, A., Shazeer, N., Parmar, N., Uszkoreit, J., Jones, L., Gomez, A.~N.,
  Kaiser, {\L}., and Polosukhin, I.
\newblock Attention is all you need.
\newblock In \emph{Advances in neural information processing systems}, pp.\
  5998--6008, 2017.

\bibitem[Veli{\v{c}}kovi{\'c} \& Blundell(2021)Veli{\v{c}}kovi{\'c} and
  Blundell]{velivckovic2021neural}
Veli{\v{c}}kovi{\'c}, P. and Blundell, C.
\newblock Neural algorithmic reasoning.
\newblock \emph{arXiv preprint arXiv:2105.02761}, 2021.

\bibitem[Veli{\v{c}}kovi{\'c} et~al.(2017)Veli{\v{c}}kovi{\'c}, Cucurull,
  Casanova, Romero, Lio, and Bengio]{velivckovic2017graph}
Veli{\v{c}}kovi{\'c}, P., Cucurull, G., Casanova, A., Romero, A., Lio, P., and
  Bengio, Y.
\newblock Graph attention networks.
\newblock \emph{arXiv preprint arXiv:1710.10903}, 2017.

\bibitem[Veli{\v{c}}kovi{\'c} et~al.(2019)Veli{\v{c}}kovi{\'c}, Ying, Padovano,
  Hadsell, and Blundell]{velivckovic2019neural}
Veli{\v{c}}kovi{\'c}, P., Ying, R., Padovano, M., Hadsell, R., and Blundell, C.
\newblock Neural execution of graph algorithms.
\newblock \emph{arXiv preprint arXiv:1910.10593}, 2019.

\bibitem[Veli{\v{c}}kovi{\'c} et~al.(2020)Veli{\v{c}}kovi{\'c}, Buesing,
  Overlan, Pascanu, Vinyals, and Blundell]{velivckovic2020pointer}
Veli{\v{c}}kovi{\'c}, P., Buesing, L., Overlan, M.~C., Pascanu, R., Vinyals,
  O., and Blundell, C.
\newblock Pointer graph networks.
\newblock \emph{arXiv preprint arXiv:2006.06380}, 2020.

\bibitem[Veli{\v{c}}kovi{\'c} et~al.(2021)Veli{\v{c}}kovi{\'c}, Bo{\v{s}}njak,
  Kipf, Lerchner, Hadsell, Pascanu, and Blundell]{velivckovic2021reasoning}
Veli{\v{c}}kovi{\'c}, P., Bo{\v{s}}njak, M., Kipf, T., Lerchner, A., Hadsell,
  R., Pascanu, R., and Blundell, C.
\newblock Reasoning-modulated representations.
\newblock \emph{arXiv preprint arXiv:2107.08881}, 2021.

\bibitem[Vinyals et~al.(2015)Vinyals, Fortunato, and
  Jaitly]{vinyals2015pointer}
Vinyals, O., Fortunato, M., and Jaitly, N.
\newblock Pointer networks.
\newblock In \emph{Advances in Neural Information Processing Systems}, pp.\
  2692--2700, 2015.

\bibitem[Williams(1964)]{williams1964algorithm}
Williams, J. W.~J.
\newblock Algorithm 232: heapsort.
\newblock \emph{Commun. ACM}, 7:\penalty0 347--348, 1964.

\bibitem[Xhonneux et~al.(2021)Xhonneux, Deac, Veli{\v{c}}kovi{\'c}, and
  Tang]{xhonneux2021transfer}
Xhonneux, L.-P., Deac, A.-I., Veli{\v{c}}kovi{\'c}, P., and Tang, J.
\newblock How to transfer algorithmic reasoning knowledge to learn new
  algorithms?
\newblock \emph{Advances in Neural Information Processing Systems}, 34, 2021.

\bibitem[Xu et~al.(2019)Xu, Li, Zhang, Du, Kawarabayashi, and
  Jegelka]{xu2019can}
Xu, K., Li, J., Zhang, M., Du, S.~S., Kawarabayashi, K.-i., and Jegelka, S.
\newblock What can neural networks reason about?
\newblock \emph{arXiv preprint arXiv:1905.13211}, 2019.

\bibitem[Xu et~al.(2020)Xu, Li, Zhang, Du, ichi Kawarabayashi, and
  Jegelka]{xu2020neural}
Xu, K., Li, J., Zhang, M., Du, S.~S., ichi Kawarabayashi, K., and Jegelka, S.
\newblock How neural networks extrapolate: From feedforward to graph neural
  networks.
\newblock \emph{arXiv preprint arXiv:2009.11848}, 2020.

\bibitem[Yan et~al.(2020)Yan, Swersky, Koutra, Ranganathan, and
  Heshemi]{yan2020neural}
Yan, Y., Swersky, K., Koutra, D., Ranganathan, P., and Heshemi, M.
\newblock Neural execution engines: Learning to execute subroutines.
\newblock \emph{arXiv preprint arXiv:2006.08084}, 2020.

\bibitem[Yehuda et~al.(2020)Yehuda, Gabel, and Schuster]{yehuda2020s}
Yehuda, G., Gabel, M., and Schuster, A.
\newblock It's not what machines can learn, it's what we cannot teach.
\newblock \emph{arXiv preprint arXiv:2002.09398}, 2020.

\bibitem[Zaheer et~al.(2017)Zaheer, Kottur, Ravanbakhsh, Poczos, Salakhutdinov,
  and Smola]{zaheer2017deep}
Zaheer, M., Kottur, S., Ravanbakhsh, S., Poczos, B., Salakhutdinov, R.~R., and
  Smola, A.~J.
\newblock Deep sets.
\newblock In \emph{Advances in neural information processing systems}, pp.\
  3391--3401, 2017.

\bibitem[Zamir et~al.(2018)Zamir, Sax, Shen, Guibas, Malik, and
  Savarese]{zamir2018taskonomy}
Zamir, A.~R., Sax, A., Shen, W., Guibas, L.~J., Malik, J., and Savarese, S.
\newblock Taskonomy: Disentangling task transfer learning.
\newblock In \emph{Proceedings of the IEEE conference on computer vision and
  pattern recognition}, pp.\  3712--3722, 2018.

\bibitem[Zaremba \& Sutskever(2014)Zaremba and Sutskever]{zaremba2014learning}
Zaremba, W. and Sutskever, I.
\newblock Learning to execute.
\newblock \emph{arXiv preprint arXiv:1410.4615}, 2014.

\end{thebibliography}

\newpage
\appendix
\onecolumn

\section{Interfacing with the CLRS benchmark}
\label{app:interface}

The CLRS benchmark is publicly hosted on GitHub: \url{https://github.com/deepmind/clrs}. All code and artifacts are released under an \textbf{Apache 2.0} license, which is highly permissive.

Within \texttt{clrs/examples/run.py}, we demonstrate an extensively configurable example script that evaluates a specific baseline on CLRS-30.

Our baselines are provided in JAX and Haiku \citep{haiku2020github}, but the dataset is generated using NumPy, making it possible to create learning pipelines in virtually any framework, including PyTorch and TensorFlow.

We will now highlight three key ways in which researchers can interface with the library.

\subsection{Evaluating a new baseline on CLRS-30}

To support a new baseline, the recommended path depends on how fundamentally different the baseline is to an encode-process-decode GNN. 

In most cases, we anticipate that only the processor network needs changing, and the remainder of the architecture can match our baselines. In this case, it is only necessary to implement the new processor network within \texttt{clrs/\_src/processors.py} and appropriately set \texttt{self.mpnn} within the \texttt{\_construct\_processor} method in \texttt{clrs/\_src/baselines.py}.

For more fundamentally different baselines, it is necessary to create a new class that extends the \texttt{Model} API (as found within \texttt{clrs/\_src/model.py}). \texttt{clrs/\_src/baselines.py} provides one example of how this can be done efficiently, for the case of our baselines.

\subsection{Modifying the data distribution of CLRS-30}

If users want to train and/or evaluate the models on different versions of the tasks given in CLRS-30, the key routines to modify are located in \texttt{clrs/\_src/samplers.py}. 

The easiest modification concerns the graph sizes and/or numbers of trajectories. They can be directly changed by modifying the \texttt{CLRS30} dictionary near the top of the file.

For more elaborate modifications, e.g. to the specific data sampling distributions, the users would need to modify and/or extend the relevant sampler class. As a guiding example, we provide a \texttt{SortingSampler} class which is convenient for generating inputs for sorting algorithms. The specific sampler used for each task is provided in the \texttt{SAMPLERS} dictionary towards the end of the file.

\subsection{Adding new algorithms to CLRS}

As the most elaborate of the three workflows, adding a new algorithm to the task suite requires following several steps, which are potentially comprehensive, depending on the complexity of the algorithm. However, the CLRS benchmark code still provides may helper routines for probing and batching that facilitate inclusion of novel algorithms. The steps are as follows:

\begin{enumerate}
    \item First, determine the input/hint/output specification of your algorithm, and include it within the \texttt{SPECS} dictionary of \texttt{clrs/\_src/specs.py}.
    \item Implement the desired algorithm in an abstractified form. Examples of this can be found throughout the \texttt{clrs/\_src/algorithms/} folder.
    \item Next, choose appropriate moments within the algorithm's execution to create probes that capture the inputs, outputs and all intermediate state (using the \texttt{probing.push} function). 
    \item Once generated, probes can be prepared using the \texttt{probing.finalize} method, and should be returned together with the algorithm output.
    \item Lastly, implement an appropriate input data sampler for your algorithm, and include it within the \texttt{SAMPLERS} dictionary within \texttt{clrs/\_src/samplers.py}.
\end{enumerate}

\section{Additional worked examples of algorithm trajectories}
\label{app:examples}

\paragraph{Matrix Chain Order}

As a representative dynamic programming algorithm, we visualise the steps of the procedure for optimising the order of multiplications in a chain of matrices, for multiplying matrices of size $(10\times 30)(30\times 5)(5\times 60)$, assuming a $O(n^3)$-time multiplication algorithm.

The algorithm proceeds by filling up an ``upper-triangular'' part of a dynamic programming matrix, where cell $[i, j]$ corresponds to the optimal number of operations when multiplying all the matrices between the $i$th and $j$th. Such an algorithm may also be represented in a ``pyramidal'' form as below:

\begin{center}
\includegraphics[width=\linewidth]{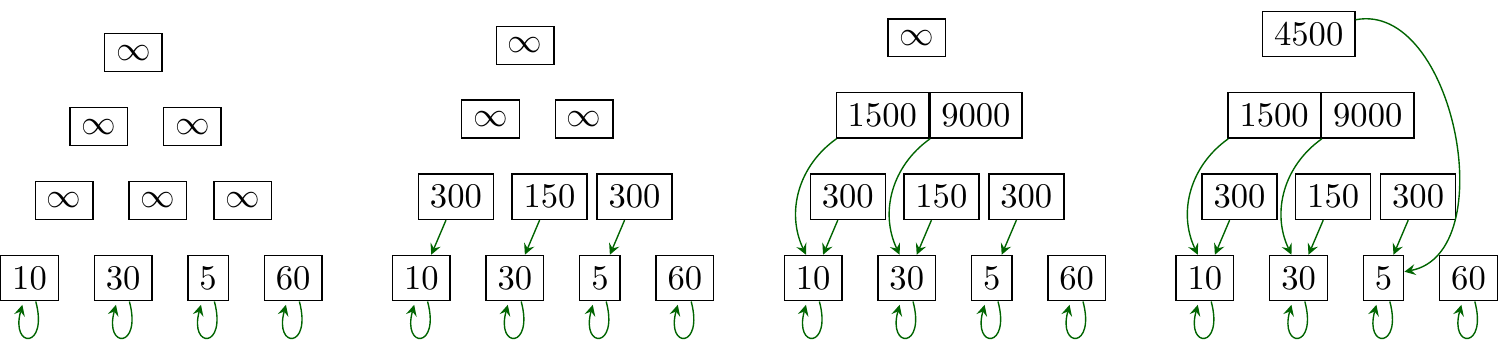}
\end{center}

Additionally, the algorithm maintains (and returns) the optimal way to recursively divide each subsequence into two (by storing the optimal dividing point, in green). Here, it is optimal to first multiply $(10\times 30)(30\times 5)$ (yielding $1,500$ operations), then multiply the remaning matrices as $(10\times 5)(5\times 60)$ (yielding $3,000$ operations; $4,500$ in total).

Note that every pointer points into one of the original $n$ input nodes (at the lowest level), and how each cell of the pyramid corresponds to a pair of input nodes (specifying the corresponding range). Therefore, rather than creating $O(n^2)$ auxiliary nodes, we instead record all relevant values above as edge scalars and edge pointers, and store nodes only for the lowest level of the pyramid. Further, whether or not a particular edge has been populated yet (the ``$\infty$'' indicator above) is stored as an additional binary flag.

\paragraph{Bellman-Ford}

As a representative graph algorithm, we visualise the steps of the Bellman-Ford algorithm for finding single-source shortest paths in a given graph. 

Initially, the source node is labelled with distance zero, and all other nodes with distance ``$\infty$'' (which, once again, is represented as a binary node hint). The algorithm then iteratively relaxes all edges as follows, until convergence is achieved:

\begin{center}
\includegraphics[width=\linewidth]{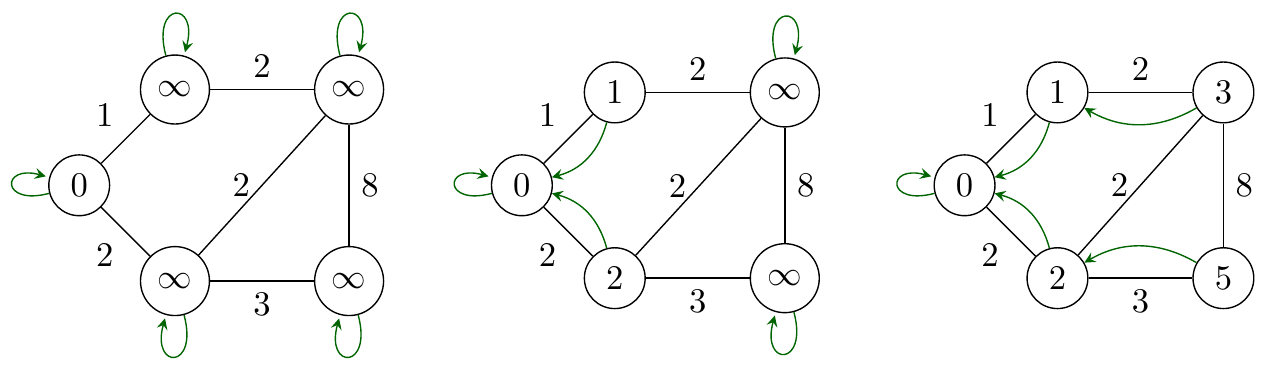}
\end{center}

Besides updating the distance values, the algorithm also maintains, and returns, the predicted shortest path tree -- for each node, a pointer to its predecessor along the optimal path from the source. By convention, the source node points to itself. These pointers are visualised in green.

\paragraph{Na\"{i}ve String Matcher}

As a representative string algorithm, we visualise the steps of the na\"{i}ve string matcher, for detecting string \texttt{"ab"} inside the string \texttt{"aab"}.

In this case, each character of the two strings is given a separate node, and three sets of indices are maintained: indicating the start of the current candidate match (in blue); and the current position being checked in both the haystack (red) and the needle (purple). The algorithm scans candidate positions left-to-right until a full match is detected for the first time.

\begin{center}
\includegraphics[width=\linewidth]{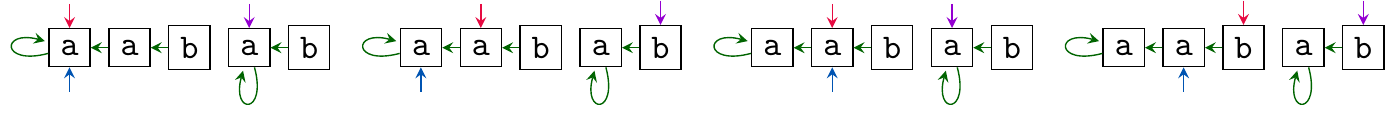}
\end{center}

Additionally, each character is tagged with its predecessor in the string (in green), and a binary flag indicating which of the two strings it belongs to (not shown here).

\section{Test results for all algorithms}
\label{app:testfull}

Test performance for all 30 algorithms in CLRS-30 may be found in Table \ref{tab:test_results_epic}. In addition, we provide a ``win-tie-loss'' metric as another way of differentiating model performance, which is less sensitive to outliers. The resulting counts are provided in Table \ref{tab:counts_results_all}, and are computed as follows:
\begin{itemize}
    \item Let $\mu_A(\mathcal{M})$ and $\sigma_A(\mathcal{M})$ be the mean and standard deviation of model $\mathcal{M}$'s test performance on algorithm $A$ (as in Table \ref{tab:test_results_epic}).
    \item We say that model $\mathcal{A}$ outperforms model $\mathcal{B}$ on algorithm $A$---denoted by $\mathcal{A}\succ_A\mathcal{B}$---if $\mu_A(\mathcal{A}) - \sigma_A(\mathcal{A}) > \mu_A(\mathcal{B})$.
    \item If $\forall \mathcal{X}\neq\mathcal{A}.\ \mathcal{A}\succ_A\mathcal{X}$, then model $\mathcal{A}$ \textbf{wins} on algorithm $A$.
    \item Otherwise, if $\exists \mathcal{X}.\ \mathcal{X}\succ_A\mathcal{A}$, then model $\mathcal{A}$ \textbf{loses} on algorithm $A$.
    \item Otherwise, model $\mathcal{A}$ is \textbf{tied} on algorithm $A$.
\end{itemize}
The win/tie/loss counts are then aggregated across all algorithms $A$ to obtain a metric for each model. As already mentioned, the details of this on a per-algorithm level are given in Table \ref{tab:counts_results_all}.

\begin{table}
    \centering
   \caption{Test performance of all models on all algorithms.}
    \begin{tabular}{lccccc}\toprule
    \textbf{Algorithm} & \textbf{Deep Sets} & \textbf{GAT} & \textbf{Memnet} & \textbf{MPNN} & \textbf{PGN} \\\midrule
Activity Selector & $ 66.09\% \pm  1.67$ & $ 73.23\% \pm  1.37$ & $ 24.10\% \pm  2.22$ & $\mathbf{ 80.66\%} \pm  3.16$ & $ 66.80\% \pm  1.62$\\
Articulation Points & $ 39.06\% \pm  4.04$ & $ 37.76\% \pm  1.62$ & $ 1.50\% \pm  0.61$ & $\mathbf{ 50.91\%} \pm  2.18$ & $ 49.53\% \pm  2.09$\\
Bellman-Ford & $ 51.33\% \pm  0.85$ & $ 87.91\% \pm  1.19$ & $ 40.04\% \pm  1.46$ & $ 92.01\% \pm  0.28$ & $\mathbf{ 92.99\%} \pm  0.34$\\
BFS & $ 98.63\% \pm  0.38$ & $ 99.04\% \pm  0.21$ & $ 43.34\% \pm  0.04$ & $\mathbf{ 99.89\% }\pm  0.05$ & $ 99.63\% \pm  0.29$\\
Binary Search & $ 47.97\% \pm  0.88$ & $ 23.50\% \pm  3.12$ & $ 14.37\% \pm  0.46$ & $ 36.83\% \pm  0.26$ & $\mathbf{ 76.95\% }\pm  0.13$\\
Bridges & $ 32.43\% \pm  2.65$ & $ 25.64\% \pm  6.60$ & $ 30.26\% \pm  0.05$ & $\mathbf{ 72.69\% \pm  4.78}$ & $ 51.42\% \pm  7.82$\\
Bubble Sort & $ 50.73\% \pm  3.24$ & $ 9.91\% \pm  1.77$ & $\mathbf{ 73.58\%} \pm  0.78$ & $ 5.27\% \pm  0.60$ & $ 6.01\% \pm  1.95$\\
DAG Shortest Paths & $ 73.21\% \pm  2.42$ & $ 81.14\% \pm  1.37$ & $ 66.15\% \pm  1.92$ & $ 96.24\% \pm  0.56$ & $\mathbf{ 96.94\%} \pm  0.16$\\
DFS & $ 7.44\% \pm  0.73$ & $ 11.78\% \pm  2.04$ & $\mathbf{ 13.36\%} \pm  1.61$ & $ 6.54\% \pm  0.51$ & $ 8.71\% \pm  0.24$\\
Dijkstra & $ 36.12\% \pm  3.10$ & $ 58.01\% \pm  0.79$ & $ 22.48\% \pm  2.39$ & $\mathbf{ 91.50\%} \pm  0.50$ & $ 83.45\% \pm  1.75$\\
Find Max. Subarray  & $ 12.48\% \pm  0.39$ & $ 24.43\% \pm  0.43$ & $ 13.05\% \pm  0.08$ & $ 20.30\% \pm  0.49$ & $\mathbf{ 65.23\%} \pm  2.56$\\
Floyd-Warshall & $ 7.22\% \pm  0.90$ & $ 16.66\% \pm  3.14$ & $ 14.17\% \pm  0.13$ & $ 26.74\% \pm  1.77$ & $\mathbf{ 28.76\%} \pm  0.51$\\
Graham Scan & $ 64.71\% \pm  2.75$ & $ 77.89\% \pm  2.70$ & $ 40.62\% \pm  2.31$ & $\mathbf{ 91.04\%} \pm  0.31$ & $ 56.87\% \pm  1.61$\\
Heapsort & $ 28.94\% \pm  12.57$ & $ 10.35\% \pm  1.83$ & $\mathbf{ 68.00\%} \pm  1.57$ & $ 10.94\% \pm  0.84$ & $ 5.27\% \pm  0.18$\\
Insertion Sort & $ 40.98\% \pm  4.65$ & $ 29.52\% \pm  1.87$ & $\mathbf{ 71.42\%} \pm  0.86$ & $ 19.81\% \pm  2.08$ & $ 44.37\% \pm  2.43$\\
Jarvis' March & $ 50.25\% \pm  0.81$ & $\mathbf{ 51.51\% }\pm  10.25$ & $ 22.99\% \pm  3.87$ & $ 34.86\% \pm  12.39$ & $ 49.19\% \pm  1.07$\\
KMP Matcher & $\mathbf{ 3.22\%} \pm  0.54$ & $ 3.03\% \pm  0.36$ & $ 1.81\% \pm  0.00$ & $ 2.49\% \pm  0.86$ & $ 2.00\% \pm  0.12$\\
LCS Length & $ 50.10\% \pm  5.25$ & $\mathbf{ 57.88\%} \pm  1.02$ & $ 49.84\% \pm  4.34$ & $ 53.23\% \pm  0.36$ & $ 56.82\% \pm  0.21$\\
Matrix Chain Order & $ 78.36\% \pm  3.58$ & $ 78.19\% \pm  3.31$ & $ 81.96\% \pm  1.03$ & $ 79.84\% \pm  1.40$ & $\mathbf{ 83.91\%} \pm  0.49$\\
Minimum & $ 80.19\% \pm  2.08$ & $ 84.20\% \pm  2.95$ & $ 86.93\% \pm  0.11$ & $ 85.34\% \pm  0.88$ & $\mathbf{ 87.71\% }\pm  0.52$\\
MST-Kruskal & $ 60.58\% \pm  4.71$ & $ 65.72\% \pm  0.99$ & $ 28.84\% \pm  0.61$ & $\mathbf{ 70.97\%} \pm  1.50$ & $ 66.96\% \pm  1.36$\\
MST-Prim & $ 12.17\% \pm  5.47$ & $ 38.20\% \pm  4.34$ & $ 10.29\% \pm  3.77$ & $\mathbf{ 69.08\%} \pm  7.56$ & $ 63.33\% \pm  0.98$\\
Na\"{i}ve String Match & $ 2.05\% \pm  0.29$ & $ 3.01\% \pm  1.20$ & $ 1.22\% \pm  0.48$ & $\mathbf{ 3.92\%} \pm  0.30$ & $ 2.08\% \pm  0.20$\\
Optimal BST & $ 69.71\% \pm  1.36$ & $ 65.49\% \pm  1.75$ & $\mathbf{ 72.03\%} \pm  1.21$ & $ 62.23\% \pm  0.44$ & $ 71.01\% \pm  1.82$\\
Quickselect & $ 3.21\% \pm  1.33$ & $\mathbf{ 4.36\%} \pm  0.95$ & $ 1.74\% \pm  0.03$ & $ 1.43\% \pm  0.69$ & $ 3.66\% \pm  0.42$\\
Quicksort & $ 37.74\% \pm  2.16$ & $ 7.60\% \pm  0.98$ & $\mathbf{ 73.10\%} \pm  0.67$ & $ 11.30\% \pm  0.10$ & $ 6.17\% \pm  0.15$\\
Segments Intersect & $ 77.29\% \pm  0.60$ & $ 90.41\% \pm  0.04$ & $ 71.81\% \pm  0.90$ & $\mathbf{ 93.44\%} \pm  0.10$ & $ 77.51\% \pm  0.75$\\
SCC & $ 17.81\% \pm  2.61$ & $ 12.70\% \pm  3.12$ & $ 16.32\% \pm  4.78$ & $\mathbf{ 24.37\%} \pm  4.88$ & $ 20.80\% \pm  0.64$\\
Task Scheduling & $ 84.84\% \pm  0.70$ & $ 84.69\% \pm  2.09$ & $ 82.74\% \pm  0.04$ & $ 84.11\% \pm  0.32$ & $\mathbf{ 84.89\%} \pm  0.91$\\
Topological Sort & $ 15.84\% \pm  3.57$ & $ 27.03\% \pm  6.92$ & $ 2.73\% \pm  0.11$ & $ 52.60\% \pm  6.24$ & $\mathbf{ 60.45\% }\pm  2.69$\\
\midrule
Overall average  & $ 43.36\%$ & $ 44.69\%$ & $ 38.03\%$ & $ 51.02\%$ & $\mathbf{ 52.31\%}$\\
\bottomrule
\end{tabular}
    \label{tab:test_results_epic}
    \end{table}

\begin{table}[ht]
    \centering
   \caption{Win/Tie/Loss counts of all models on all algorithms. Legend: {\bf W}: win, {\em T}: tie, L: loss.}
    \begin{tabular}{lccccc} \toprule
    \textbf{Algorithm} & \textbf{Deep Sets} & \textbf{GAT} & \textbf{Memnet} & \textbf{MPNN} & \textbf{PGN} \\\midrule
Activity Selector & L & L & L & {\bf W} & L\\
Articulation Points & L & L & L & {\em T} & {\em T}\\
Bellman-Ford & L & L & L & L & {\bf W}\\
BFS & L & L & L & {\bf W} & L\\
Binary Search & L & L & L & L & {\bf W}\\
Bridges & L & L & L & {\bf W} & L\\
Bubble Sort & L & L & {\bf W} & L & L\\
DAG Shortest Paths & L & L & L & L & {\bf W}\\
DFS & L & {\em T} & {\em T} & L & L\\
Dijkstra & L & L & L & {\bf W} & L\\
Find Max. Subarray  & L & L & L & L & {\bf W}\\
Floyd-Warshall & L & L & L & L & {\bf W}\\
Graham Scan & L & L & L & {\bf W} & L\\
Heapsort & L & L & {\bf W} & L & L\\
Insertion Sort & L & L & {\bf W} & L & L\\
Jarvis' March & {\em T} & {\em T} & L & L & L\\
KMP Matcher & {\em T} & {\em T} & L & L & L\\
LCS Length & L & {\bf W} & L & L & L\\
Matrix Chain Order & L & L & L & L & {\bf W}\\
Minimum & L & L & L & L & {\bf W}\\
MST-Kruskal & L & L & L & {\bf W} & L\\
MST-Prim & L & L & L & {\em T} & {\em T}\\
Na\"{i}ve String Match & L & L & L & {\bf W} & L\\
Optimal BST & L & L & {\em T} & L & {\em T}\\
Quickselect & L & {\em T} & L & L & {\em T}\\
Quicksort & L & L & {\bf W} & L & L\\
Segments Intersect & L & L & L & {\bf W} & L\\
SCC & L & L & L & {\em T} & {\em T}\\
Task Scheduling & {\em T} & {\em T} & L & L & {\em T}\\
Topological Sort & L & L & L & L & {\bf W}\\
\midrule
Overall counts  & 0/3/27 & 1/5/24 & 4/2/24 & 8/3/19 & {\bf 8/6/16}\\
\bottomrule
\end{tabular}
    \label{tab:counts_results_all}
    \end{table}
    
\section{Validation results individual plots}
\label{app:vertical}

\begin{figure}
\centering
\includegraphics[width=\textwidth]{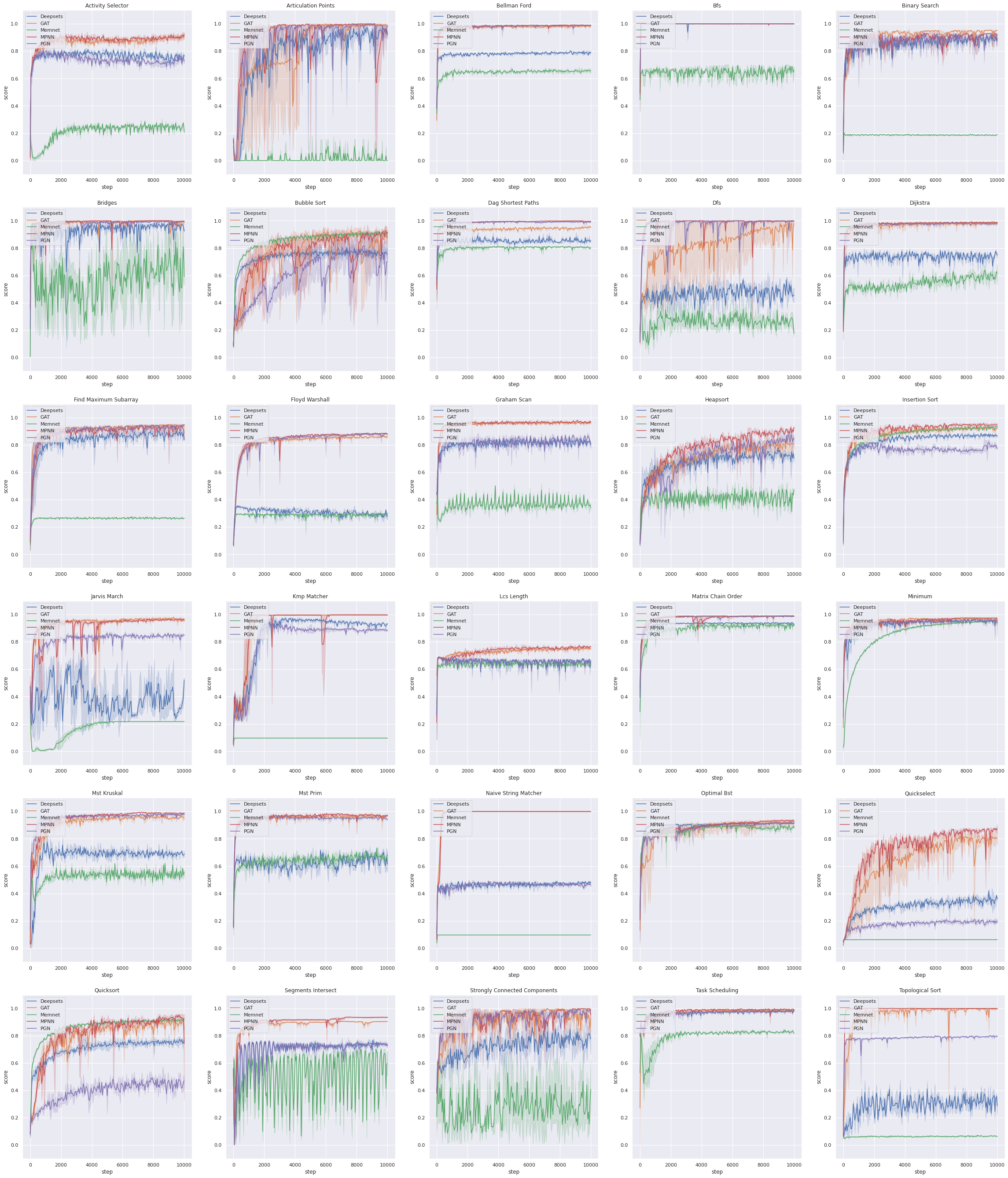}
\caption{Validation results on all 30 algorithms in CLRS-30, averaged over three seeds.}
\label{fig:plots_2_epic}
\end{figure}

Validation performance for all 30 algorithms in CLRS-30 may be found in Figure \ref{fig:plots_2_epic}. For convenience, we also report the early-stopped validation performance in Table \ref{tab:val_results_all}.

\begin{table}[ht]
    \centering

   \caption{Early-stopped validation results of all models on all algorithms.}
    \begin{tabular}{lccccc} \toprule
    \textbf{Algorithm} & \textbf{Deep Sets} & \textbf{GAT} & \textbf{Memnet} & \textbf{MPNN} & \textbf{PGN} \\\midrule
Activity Selector & $ 83.50\% \pm  0.17$ & $ 92.40\% \pm  0.50$ & $ 34.59\% \pm  2.15$ & $\mathbf{ 93.89\%} \pm  0.39$ & $ 82.26\% \pm  0.19$\\
Articulation Points & $ 99.63\% \pm  0.31$ & $\mathbf{ 100.00\%} \pm  0.00$ & $ 16.84\% \pm  1.03$ & $\mathbf{ 100.00\%} \pm  0.00$ & $\mathbf{ 100.00\%} \pm  0.00$\\
Bellman-Ford & $ 81.12\% \pm  0.14$ & $ 99.28\% \pm  0.14$ & $ 68.75\% \pm  0.42$ & $\mathbf{ 99.48\%} \pm  0.05$ & $ 99.35\% \pm  0.05$\\
BFS & $\mathbf{ 100.00\%} \pm  0.00$ & $\mathbf{ 100.00\%} \pm  0.00$ & $ 70.70\% \pm  0.09$ & $\mathbf{ 100.00\%} \pm  0.00$ & $\mathbf{ 100.00\%} \pm  0.00$\\
Binary Search & $ 93.34\% \pm  0.41$ & $\mathbf{ 95.72\%} \pm  0.17$ & $ 20.33\% \pm  0.28$ & $ 94.19\% \pm  0.12$ & $ 94.17\% \pm  0.08$\\
Bridges & $ 99.36\% \pm  0.05$ & $\mathbf{ 100.00\%} \pm  0.00$ & $ 96.46\% \pm  1.13$ & $\mathbf{ 100.00\%} \pm  0.00$ & $\mathbf{ 100.00\%} \pm  0.00$\\
Bubble Sort & $ 81.51\% \pm  1.02$ & $\mathbf{ 95.44\% }\pm  1.01$ & $ 92.64\% \pm  0.14$ & $ 94.53\% \pm  1.84$ & $ 87.17\% \pm  5.46$\\
DAG Shortest Paths & $ 92.25\% \pm  0.28$ & $ 96.81\% \pm  0.05$ & $ 81.90\% \pm  0.05$ & $\mathbf{ 99.93\%} \pm  0.05$ & $ 99.80\% \pm  0.00$\\
DFS & $ 62.76\% \pm  1.26$ & $ 99.22\% \pm  0.64$ & $ 47.72\% \pm  0.45$ & $\mathbf{ 100.00\% }\pm  0.00$ & $\mathbf{ 100.00\%} \pm  0.00$\\
Dijkstra & $ 80.34\% \pm  0.42$ & $ 99.22\% \pm  0.40$ & $ 67.38\% \pm  0.70$ & $\mathbf{ 99.67\%} \pm  0.14$ & $ 99.28\% \pm  0.05$\\
Find Max. Subarray  & $ 91.41\% \pm  0.22$ & $ 95.00\% \pm  0.32$ & $ 27.91\% \pm  0.08$ & $ 95.13\% \pm  0.37$ & $\mathbf{ 95.30\%} \pm  0.16$\\
Floyd-Warshall & $ 35.79\% \pm  0.04$ & $ 87.28\% \pm  0.09$ & $ 31.29\% \pm  0.04$ & $\mathbf{ 89.14\% }\pm  0.03$ & $ 88.70\% \pm  0.15$\\
Graham Scan & $ 87.66\% \pm  0.24$ & $ 97.85\% \pm  0.11$ & $ 53.53\% \pm  1.58$ & $\mathbf{ 98.45\%} \pm  0.15$ & $ 89.06\% \pm  0.27$\\
Heapsort & $ 81.84\% \pm  0.33$ & $ 87.24\% \pm  2.23$ & $ 54.04\% \pm  0.28$ & $\mathbf{ 94.27\%} \pm  0.11$ & $ 90.36\% \pm  0.67$\\
Insertion Sort & $ 89.58\% \pm  0.28$ & $ 95.18\% \pm  0.58$ & $ 94.40\% \pm  0.14$ & $\mathbf{ 96.74\%} \pm  0.19$ & $ 84.57\% \pm  0.82$\\
Jarvis' March & $ 72.82\% \pm  0.42$ & $\mathbf{ 98.38\%} \pm  0.16$ & $ 37.92\% \pm  6.61$ & $ 97.94\% \pm  0.25$ & $ 88.34\% \pm  0.36$\\
KMP Matcher & $ 98.03\% \pm  0.21$ & $ 99.76\% \pm  0.08$ & $ 9.67\% \pm  0.00$ & $\mathbf{ 99.87\%} \pm  0.05$ & $ 94.14\% \pm  0.99$\\
LCS Length & $ 69.24\% \pm  0.36$ & $ 77.00\% \pm  0.19$ & $ 67.69\% \pm  0.24$ & $\mathbf{ 77.88\%} \pm  0.42$ & $ 69.19\% \pm  0.04$\\
Matrix Chain Order & $ 94.46\% \pm  0.02$ & $\mathbf{ 99.37\%} \pm  0.03$ & $ 93.91\% \pm  0.10$ & $ 99.12\% \pm  0.04$ & $ 99.21\% \pm  0.03$\\
Minimum & $ 97.59\% \pm  0.11$ & $\mathbf{ 97.74\%} \pm  0.21$ & $ 95.56\% \pm  0.10$ & $ 97.64\% \pm  0.05$ & $ 97.07\% \pm  0.14$\\
MST-Kruskal & $ 83.79\% \pm  2.01$ & $ 97.93\% \pm  0.25$ & $ 64.65\% \pm  0.95$ & $\mathbf{ 99.71\%} \pm  0.17$ & $ 99.12\% \pm  0.08$\\
MST-Prim & $ 74.61\% \pm  0.32$ & $ 98.37\% \pm  0.14$ & $ 74.09\% \pm  0.28$ & $\mathbf{ 99.02\%} \pm  0.09$ & $ 97.79\% \pm  0.14$\\
Na\"{i}ve String Match & $ 49.80\% \pm  0.15$ & $\mathbf{ 100.00\%} \pm  0.00$ & $ 9.91\% \pm  0.20$ & $\mathbf{ 100.00\%} \pm  0.00$ & $ 50.33\% \pm  0.08$\\
Optimal BST & $ 92.02\% \pm  0.14$ & $ 93.30\% \pm  0.49$ & $ 90.86\% \pm  0.40$ & $\mathbf{ 93.88\%} \pm  0.11$ & $ 93.20\% \pm  0.27$\\
Quickselect & $ 42.30\% \pm  0.92$ & $ 83.82\% \pm  1.86$ & $ 6.56\% \pm  0.25$ & $\mathbf{ 88.74\%} \pm  0.78$ & $ 54.02\% \pm  0.17$\\
Quicksort & $ 79.69\% \pm  1.12$ & $ 92.97\% \pm  0.40$ & $ 93.16\% \pm  0.24$ & $\mathbf{ 95.70\%} \pm  0.40$ & $ 54.30\% \pm  1.42$\\
Segments Intersect & $ 77.49\% \pm  0.12$ & $ 90.82\% \pm  0.16$ & $ 71.57\% \pm  1.08$ & $\mathbf{ 93.84\%} \pm  0.20$ & $ 78.32\% \pm  0.18$\\
SCC & $ 89.52\% \pm  1.23$ & $\mathbf{ 100.00\%} \pm  0.00$ & $ 70.57\% \pm  1.43$ & $\mathbf{ 100.00\%} \pm  0.00$ & $ 99.93\% \pm  0.05$\\
Task Scheduling & $ 99.16\% \pm  0.04$ & $ 99.80\% \pm  0.04$ & $ 84.80\% \pm  0.09$ & $\mathbf{ 100.00\%} \pm  0.00$ & $ 99.06\% \pm  0.08$\\
Topological Sort & $ 47.23\% \pm  0.81$ & $\mathbf{ 100.00\%} \pm  0.00$ & $ 8.30\% \pm  0.50$ & $\mathbf{ 100.00\%} \pm  0.00$ & $\mathbf{ 100.00\%} \pm  0.00$\\
\midrule
Overall average  & $ 80.93\%$ & $ 95.66\%$ & $ 57.92\%$ & $\mathbf{ 96.63\%}$ & $ 89.47\%$\\
\bottomrule
\end{tabular}
    \label{tab:val_results_all}
    \end{table}

\end{document}